\documentclass{article}

\usepackage[nonatbib,final]{neurips_2023}
\usepackage{lipsum}
\usepackage{comment}

\usepackage{graphicx}
\usepackage{amsmath}
\usepackage{amssymb}
\usepackage{comment}
\usepackage{lipsum}
\usepackage{booktabs}
\usepackage{multirow}
\usepackage{floatrow}

\usepackage[utf8]{inputenc} 
\usepackage[T1]{fontenc}    
\usepackage{hyperref}       
\usepackage{url}            
\usepackage{booktabs}       
\usepackage{amsfonts}       
\usepackage{nicefrac}       
\usepackage{microtype}      
\usepackage[numbers,sort&compress]{natbib}
\usepackage{enumitem}
\usepackage{color}
\usepackage{colortbl}
\usepackage[table]{xcolor}
\usepackage{wrapfig}
\usepackage{makecell}
\usepackage{float}
\usepackage[final]{pdfpages}

\floatstyle{plaintop}
\restylefloat{table}

\newfloatcommand{capbtabbox}{table}[][\FBwidth]

\definecolor{yellow}{rgb}{1,1, 0.6}
\definecolor{lightyellow}{rgb}{1,1, 0.8}
\definecolor{orange}{rgb}{1, 0.8, 0.6}
\definecolor{red}{rgb}{1, 0.6, 0.6}

\title{Transient Neural Radiance Fields\\for Lidar View Synthesis and 3D Reconstruction}

\author{%
  Anagh Malik$^{1,2}$ \\
  {anagh@cs.toronto.edu}
  \And
  Parsa Mirdehghan$^{1,2}$\\
  {parsa@cs.toronto.edu}
  \And
    Sotiris Nousias$^{1}$ \\
  {sotiris@cs.toronto.edu}
  \And
Kiriakos N. Kutulakos$^{1,2}$ \\
  {kyros@cs.toronto.edu}
  \And
David B. Lindell$^{1,2}$ \\
  {lindell@cs.toronto.edu}
\AND
  {\normalfont $^1$University of Toronto \hspace{10pt}}
  {\normalfont $^2$Vector Institute \hspace{10pt}} \\
  \url{anaghmalik.com/TransientNeRF}
}

\begin{document}

\maketitle

\begin{abstract}
Neural radiance fields (NeRFs) have become a ubiquitous tool for modeling scene appearance and geometry from multiview imagery. Recent work has also begun to explore how to use additional supervision from lidar or depth sensor measurements in the NeRF framework. However, previous lidar-supervised NeRFs focus on rendering conventional camera imagery and use lidar-derived point cloud data as auxiliary supervision; thus, they fail to incorporate the underlying image formation model of the lidar. Here, we propose a novel method for rendering \textit{transient} NeRFs that take as input the raw, time-resolved photon count histograms measured by a single-photon lidar system, and we seek to render such histograms from novel views. Different from conventional NeRFs, the approach relies on a time-resolved version of the volume rendering equation to render the lidar measurements and capture transient light transport phenomena at picosecond timescales. We evaluate our method on a first-of-its-kind dataset of simulated and captured transient multiview scans from a prototype single-photon lidar. Overall, our work brings NeRFs to a new dimension of imaging at transient timescales, newly enabling rendering of transient imagery from novel views. Additionally, we show that our approach recovers improved geometry and conventional appearance compared to point cloud-based supervision when training on few input viewpoints. Transient NeRFs may be especially useful for applications which seek to simulate raw lidar measurements for downstream tasks in autonomous driving, robotics, and remote sensing.

\end{abstract}


\section{Introduction}

The ability to sense and reconstruct 3D appearance and geometry is critical to applications in vision, graphics, and beyond. 
Lidar sensors~\cite{seitz2011single} are of particular interest for this task due to their high sensitivity to arriving photons and their extremely high temporal resolution; as such, they are being deployed in systems for 3D imaging in smart phone cameras~\cite{chugunov2022implicit}, autonomous driving, and remote sensing~\cite{schwarz2010mapping}.
Recent work has also begun to explore how additional supervision from lidar~\cite{rematas2022urban} or depth sensor measurements~\cite{attal2021torf} can be incorporated into the NeRF framework to improve novel view synthesis and 3D reconstruction.
Existing NeRF-based methods that use lidar~\cite{rematas2022urban} are limited to rendering \textit{conventional RGB images}, and use lidar point clouds (i.e., pre-processed lidar measurements) as auxiliary supervision rather than rendering the raw data that lidar systems actually collect.
Specifically, lidars capture \textit{transient images}---time-resolved picosecond- or nanosecond-scale measurements of a pulse of light travelling to a scene point and back. 
We consider the problem of how to synthesize such transients from novel viewpoints.
In particular, we seek a method that takes as input and renders transients in the form of time-resolved photon count histograms captured by a single-photon lidar system\footnote{Single-photon lidars are closely related to conventional lidar systems based on avalanche photodiodes~\cite{campbell2004recent}, but they have improved sensitivity and timing resolution (discussed in Section~\ref{sec:related_work}); other types of coherent lidar systems~\cite{poulton2017coherent} are outside the scope considered here.}~\cite{rapp2020advances}.
Lidar view synthesis may be useful for applications that seek to simulate raw lidar measurements for downstream tasks, including autonomous driving, robotics, remote sensing, and virtual reality.

The acquisition and reconstruction of transient measurements has been studied across various different sensing modalities, including holography~\cite{abramson1978light}, photonic mixer devices~\cite{heide2013low,kadambi2013coded} streak cameras~\cite{velten2013femto}, and single-photon detectors (SPADs)~\cite{o2017reconstructing,kirmani2009looking}. 

In the context of SPADs and single-photon lidar, a transient is measured by repeatedly illuminating a point with pulses of light and accumulating the individual photon arrival times into a time-resolved histogram.  
After capturing such histograms for each point in a scene, one can exploit their rich spatio-temporal structure for scene reconstruction~\cite{lindell2018single,peng2020photon}, to uncover statistical properties of captured photons~\cite{rapp2017few,rapp2021high}, and to reveal the temporal profile of the laser pulse used to probe the scene (knowledge of which can significantly improve depth resolution~\cite{heide2018sub,rapp2020dithered}).
These properties motivate transients as a representation and their synthesis from novel views.
While existing methods have explored multiview lidar reconstruction~\cite{hess2016real,zou2021comparative,li2019dense,leotta2019urban,lionar2021neuralblox}, they exclusively use point cloud data, and do not tackle lidar view synthesis.

\begin{figure}[t]
    \includegraphics[]{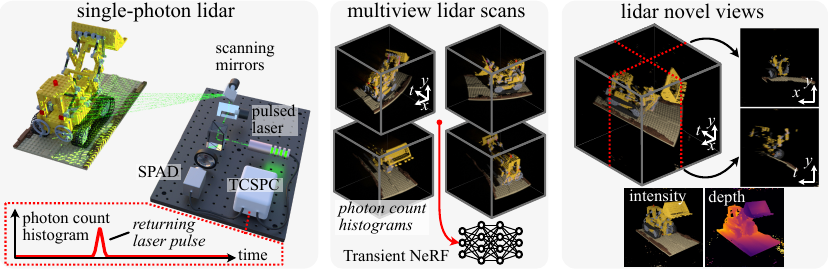}
    \caption{Overview of transient neural radiance fields (Transient NeRFs). Measurements from a single-photon lidar are captured using a single-photon avalanche diode (SPAD), pulsed laser, scanning mirrors, and a time-correlated single photon counter (TCSPC). The lidar scans, consisting of a 2D array of photon count histograms (visualized with maximum-intensity projection), are captured from multiple viewpoints and used to optimize the transient NeRF. After training, we render novel views of time-resolved lidar measurements ($x$--$y$ and $x$--$t$ slices are indicated by the dotted red lines), and we also convert the rendered data into intensity and depth maps.}
    \label{fig:teaser}
\end{figure}

Recently, a number of NeRF-based methods for 3D scene modeling have also been proposed to incorporate point cloud data (e.g., from lidar or structure from motion)~\cite{deng2022depth,roessle2022dense} or information from time-of-flight sensors~\cite{attal2021torf}.
Again, these methods focus on synthesizing conventional RGB images or depth maps, while our approach synthesizes transient images.  
Another class of methods combines NeRFs with single-photon lidar data for non-line-of-sight imaging~\cite{faccio2020non}; however, they focus on a very different inverse problem and scene parameterization~\cite{shen2021non}, and do not aim to perform novel view synthesis of lidar data as we do.

Our approach, illustrated in Fig.~\ref{fig:teaser}, extends neural radiance fields to be compatible with a statistical model of time-resolved measurements captured by a single-photon lidar system.
The method takes as input multiview scans from a single-photon lidar and, after training, enables rendering lidar measurements from novel views.
Moreover, accurate depth maps or intensity images can also be rendered from the learned representation.

In summary, we make the following contributions.
\begin{itemize}[noitemsep,topsep=0pt,leftmargin=*]
    \item We develop a novel time-resolved volumetric image formation model for single-photon lidar and introduce transient neural radiance fields for lidar view synthesis and 3D reconstruction. 
    \item We assemble a first-of-its-kind dataset of simulated and captured transient multiview scans, constructed using a prototype multiview single-photon lidar system.
    \item We use the dataset to demonstrate new capabilities in transient view synthesis and state-of-the-art results on 3D reconstruction and appearance modeling from few (2--5) single-photon lidar scans of a scene.
\end{itemize}

\section{Related work}
\label{sec:related_work}

Our work ties together threads from multiple areas of previous research, including methods for imaging with single-photon sensors, and NeRF-based pipelines that leverage 3D information to improve reconstruction quality.
Our implementation also builds on recent frameworks that improve the computational efficiency of NeRF training~\cite{muller2022instant,li2022nerfacc}.

\paragraph{Active single-photon imaging.}
Single-photon sensors output precise timestamps corresponding to the arrival times of individual detected photons.
The most common type of single-photon sensor is the single-photon avalanche diode (SPAD). SPADs are based on the widely-available CMOS technology~\cite{zappa2007principles} (which we consider in this work), but other technologies such as superconducting nanowire single-photon detectors~\cite{natarajan2012superconducting} and silicon photomultipliers~\cite{buzhan2003silicon}, offer different tradeoffs in terms of sensitivity, temporal resolution, and cost. 

In active imaging scenarios, pulsed light sources are paired with single-photon sensors to estimate the depth or reflectance of a scene by applying computational algorithms to the captured photon timestamps~\cite{shin2016photon,tachella2019real,heide2018sub}.
The extreme temporal resolution of these sensors also enables direct capture of interactions of light with a scene at picosecond timescales~\cite{gariepy2015single,lindell2018towards}, and by modeling and inverting the time-resolved scattering of light, single-photon sensors can be used to see around corners~\cite{faccio2020non,rapp2020seeing,xin2019theory,chen2020learned} or through scattering media~\cite{lindell2020three,zhao2021high,tobin2019three}.
The extreme sensitivity of single-photon sensors has made them an attractive technology for autonomous navigation~\cite{rapp2020advances}, and accurate depth acquisition from mobile phones~\cite{chugunov2022implicit}.

Our approach differs significantly from all the previous work in that we investigate, for the first time, the problem of lidar view synthesis and multi-view 3D reconstruction in the single-photon lidar regime.
We introduce the framework of transient NeRFs for this task and jointly optimize a representation of scene geometry and appearance that is consistent with captured photon timestamps across all input views.

Finally, we note that while commercial lidars based on SPADs or avalanche photodiodes capture photon count histograms or time-resolved intensity, they typically pre-process these raw measurements to point cloud format before output.  
Hence, the raw lidar data that we use may not have been widely available to previous methods; we have publicly released our dataset of multiview photon count histograms on the project webpage to help alleviate this challenge.  


\paragraph{3D-informed neural radiance fields.}
A number of recent techniques for multiview reconstruction using NeRFs leverage additional geometric information (sparse point clouds from lidar~\cite{rematas2022urban} or structure from motion~\cite{deng2022depth,roessle2022dense}) to improve the reconstruction quality or reduce the number of required input viewpoints. 
Similar benefits can be obtained by combining volume rendering with monocular depth estimators~\cite{yumonosdf}, or using data from time-of-flight sensors~\cite{attal2021torf}. 
Other methods investigate the problem of view synthesis from few input images but leverage appearance priors instead of explicit depth supervision~\cite{niemeyer2022regnerf,yu2021pixelnerf,sinha2022sparsepose}.
In contrast to the proposed approach, all of these methods focus on reconstructing images or depth maps rather than transients.

\section{Transient Neural Radiance Fields}

We describe a mathematical model for transient measurements captured using single-photon lidar and propose a time-resolved volume rendering formulation compatible with neural radiance fields.

\subsection{Image Formation Model}

\begin{figure*}[t]
    \centering
    \includegraphics[]{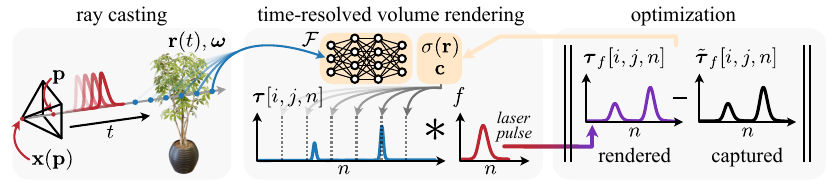}
    \caption{Rendering transient neural radiance fields. We cast rays through a volume and retrieve the density and color at each point using a neural representation~\cite{muller2022instant}. A time-resolved measurement is constructed using volume rendering (Equation~\ref{eq:nerf}), and we bin the radiance contributions into an array based on distance along the ray. The result is convolved with the impulse response of the lidar (which incorporates the shape of the laser pulse), and we supervise the neural representation based on the difference between the rendered and captured transient measurements.}
    \label{fig:method}
\end{figure*}

Consider that a laser pulse illuminates a point in a scene that is imaged onto a sensor at position $\mathbf{p} \in \mathbb{R}^2$ (see Fig.~\ref{fig:method}).
Assume light from the laser pulse propagates to a surface and back to $\mathbf{p}$ along the same path described by a ray $\mathbf{r}(t)$, where $t$ indicates propagation time. 
The forward path along the ray is given as $\mathbf{r}(t) = \mathbf{x}(\mathbf{p}) + tc\, \boldsymbol{\omega}(\mathbf{p})$, where  $\mathbf{x}(\mathbf{p})\in\mathbb{R}^3$ is the ray origin, $\boldsymbol{\omega}(\mathbf{p})\in\mathbb{S}^2$ is the ray direction which maps to $\mathbf{p}$, and $c$ is the speed of light.
Now, let $f(t)$ denote the temporal impulse response of the lidar (including the temporal profile of the laser pulse and the sensor jitter), and let $\alpha(\mathbf{p})$ incorporate reflectance and radiometric falloff factors~\cite{rapp2017few} of the illuminated point at distance $z(\mathbf{p})$ from $\mathbf{x}(\mathbf{p})$. 
Then, assuming single-bounce light transport, the photon arrival rate incident on the sensor from the laser pulse is given as 
\begin{equation}
    \boldsymbol{\lambda}[i, j, n] = \int_{\mathcal{P}_{i, j}} \int_{\mathcal{T}_n} \alpha(\mathbf{p}) \,f \left(t - \frac{2z(\mathbf{p})}{c}\right) \,\mathrm{d}t\, \mathrm{d}\mathbf{p},
    \label{eq:lidar}
\end{equation}
where $\mathcal{T}_n$ and $\mathcal{P}_{i,j}$ indicate the temporal and spatial discretization intervals for the time bin $n$ and pixel $i, j$, respectively. 
The term $2z/c$ gives the time delay for light to propagate to a point at distance $z$ and back.

Now, we can describe the measured transient, or the number of photon detections captured by a SPAD~\cite{rapp2017few}, as 
\begin{align}
    \boldsymbol{\widetilde{\tau}}[i, j, n] &\sim \textsc{Poisson} \left(N \eta\, \boldsymbol{\lambda}[i, j, n] + B\right),\quad
    B = N(\eta A[i, j] + D),
    \label{eq:histogram}
\end{align}
where $N$ indicates the number of laser pulses per pixel, $\eta \in (0, 1)$ is the detection efficiency of the sensor, and $B$ is the total number of background (non-laser pulse) detections. Background detections in turn depend on $A$, the average ambient photon rate at pixel $[i, j]$, and $D$, number of false detections produced by the sensor per laser pulse period, also known as the dark count rate.
When the number of detected photons is far fewer than the number of laser pulses, SPAD measurements can be modeled according to a Poisson process~\cite{rapp2017few} where the arrival rate function varies across space and time.
This model is appropriate for our measurements, which have relatively low flux ($<5\%$ detection probability per emitted laser pulse)~\cite{heide2018sub}.
The resulting measurements $\boldsymbol{\widetilde{\tau}}[i, j, n]$ represent a noisy histogram of photon counts collected at pixel $[i, j]$ at time bin $n$.

\subsection{Time-Resolved Volume Rendering}
Using the measurements $\boldsymbol{\widetilde{\tau}}$, we wish to optimize a representation of the appearance and geometry of the scene.
To this end, we propose a time-resolved version of the volume rendering equation used in NeRF~\cite{mildenhall2021nerf,kajiya1984ray}.
Specifically, we model clean (i.e., without Poisson noise) time-resolved histograms $\boldsymbol{\tau}[i, j, n]$ as (writing $\mathbf{r}(t)$ as $\mathbf{r}$ for brevity) 
\begin{align}
    \boldsymbol{\tau}[i, j, n] &= \int_{\mathcal{P}_{i, j}} \int_{\mathcal{T}_n} (tc)^{-2}\,T(t)^2\sigma(\mathbf{r})\mathbf{c}(\mathbf{r}, \boldsymbol{\omega}) \,\mathrm{d}t\, \mathrm{d}\mathbf{p},\nonumber \\ 
                        \text{where} \,\,\, T(t) &= \exp\left( -\int_{t_0}^{t}\sigma(\mathbf{r})\,\mathrm{d}s\right).
    \label{eq:nerf}
\end{align}
We denote by $\mathbf{c}$ the radiance of light scattered at a point $\mathbf{r}(t)$ in the direction $\boldsymbol{\omega}$, and $\sigma$ represents the volume density or the differential probability of ray termination at $\mathbf{r}(t)$.
Finally, $T(t)$ is the transmittance from a distance $t_0$ along the ray to $t$, and this term is squared to reflect the two-way propagation of light~\cite{attal2021torf}. We additionally explicitly account for the inverse-square falloff of intensity, through the term  $(tc)^{-2}$.
The definite integrals are evaluated over the extent of time bin $n$, $\mathcal{T}_n = [t_{n-1}, t_n]$, and over $\mathbf{p}$ within the area of pixel $[i, j]$ as in Equation~\ref{eq:lidar}.
Note that in practice, we calculate Equation~\ref{eq:lidar} using the discretization scheme of Max~\cite{max1995optical} used by Mildenhall et al.~\cite{mildenhall2021nerf}. 

Finally, to account for the temporal spread of the laser pulse and sensor jitter, we convolve the estimated transient with the calibrated impulse response of the lidar system $f$ to obtain

\begin{align}
\boldsymbol{\tau}_f = f \ast \boldsymbol{\tau}.
\end{align}

Without this step, the volumetric model of Equation \ref{eq:nerf} does not match the raw data from the lidar system and tends to produce thick clouds of density around surfaces to compensate for this mismatch.

\subsection{Reconstruction}

To reconstruct transient NeRFs, we use lidar measurements $\boldsymbol{\widetilde{\tau}}^{(k)}[i, j, n]$ of a scene captured from $0 \leq k \leq K-1$ different viewpoints.
We parameterize transient NeRF using a neural network $\mathcal{F}$ consisting of a hash grid of features and a multi-layer perceptron decoder~\cite{muller2022instant}. The network takes as input a coordinate and viewing direction, and outputs radiance and density, $\mathcal{F}(\mathbf{r}(t), \boldsymbol{\omega}) = \mathbf{c}, \sigma$.
We use these outputs to render transients (see Fig.~\ref{fig:method}). The model is optimized to minimize the difference between the rendered transient and measured photon count histograms.
We also introduce a modified loss function to account for the high dynamic range of lidar measurements, and we propose a space carving regularization penalty to help mitigate convergence to local minima in the optimization.

\paragraph{HDR-Informed loss function.}
Measurements from a single photon sensor can have a dynamic range that spans multiple orders of magnitude, with each pixel recording from zero to thousands of photons. 
We find that applying two exponential functions to the radiance preactivations (1) enforces non-negativity and (2) improves the dynamic range of the network output.
Thus, we have $\mathbf{c} = \exp(\exp(\mathbf{\hat{c}}))-1$, where the network preactivations are given by $\mathbf{\hat{c}}$.
Following Muller et al.~\cite{muller2022instant} the network also predicts density in log space.

After time-resolved volume rendering using Equation~\ref{eq:nerf}, we apply a loss function in log space to prevent the brightest regions from dominating the loss~\cite{mildenhall2022rawnerf}.
The loss function is given as
\begin{align}
\mathcal{L}_{\boldsymbol{\tau}} = \sum_{k, i, j, n} \rVert\ln(\boldsymbol{\widetilde{\tau}}^{(k)}[i, j, n]+1) - \ln(\boldsymbol{\tau}_f^{(k)}[i, j, n]+1)\lVert_1,
\end{align}
where the sum is over all images, pixels, and time bins.

\paragraph{Space carving regularization.}
We find that using the above loss function alone results in spurious patches of density in front of dark surfaces in a scene. 
Here, the network can predict bright values on the surface itself, but darkens the corresponding values of $\boldsymbol{\tau}_f$ by placing additional spurious density values along the ray. 
Since the network can predict the radiance of the density to be zero at these points, the predicted transients $\boldsymbol{\tau}_f$ can be entirely consistent with the measured transients $\boldsymbol{\widetilde{\tau}}$, but with incorrect geometry.
To address this, we introduce a space carving regularization
\begin{align}
    \mathcal{L}_{\text{sc}} = \sum\limits_{\substack{k,i, j, n \\ \boldsymbol{\widetilde{\tau}}^{(k)}[i, j, n] < B}} \int_{\mathcal{P}_{i, j}} \int_{\mathcal{T}_n} T(t)\sigma(\mathbf{r}) \,\mathrm{d}t\, \mathrm{d}\mathbf{p}.
\end{align}
This function penalizes any density along a ray at locations where the corresponding measured transient values are less than the expected background level $B$. 
This effectively forces space to be empty (i.e., zero density) at regions where the measurements do not indicate the presence of a surface.

The complete loss function used for training is then given as 
\begin{equation}
    \mathcal{L} = \mathcal{L}_{\boldsymbol{\tau}} + \lambda_{\text{sc}} \mathcal{L}_{\text{sc}},
\end{equation}
where $\lambda_{\text{sc}}$ controls the strength of the space carving regularization.

\subsection{Implementation Details}

Our implementation is based on the NerfAcc \cite{li2022nerfacc} version of Instant-NGP \cite{muller2022instant}, which we extend to incorporate our time-resolved volume rendering equation.
In particular, we extend the framework to output time-resolved transient measurements, to account for the pixel footprint, and to estimate depth.

\paragraph{Pixel footprint.}
Captured photon count histograms exhibit multiple peaks where the finite beam width of the laser passes over depth discontinuities. 
To account for this phenomenon we use a truncated Gaussian distribution to model the spatial footprint of the laser spot and SPAD sensor projected onto the scene.
Specifically, we sample rays in the range of 4 standard deviations of the pixel center, weighting their contribution to the rendered transient by the corresponding Gaussian probability density function value (after rendering a histogram).  
We set the standard deviation of the Gaussian to 0.15 pixels for the simulated dataset and 0.10 pixels for the captured dataset.

\paragraph{Depth.}
To estimate depth we find the distance along each ray that results in the maximum probability of ray termination: $\mathrm{argmax}_t\; T(t)\sigma(t)$.
Note that when integrating over the pixel footprint at occlusion boundaries, multiple local extrema can occur, and so taking the highest peak results in a single depth estimate without floating pixel artifacts. 

\paragraph{Network optimization.}
We optimize the network using the Adam optimizer~\cite{kingma2015adam}, a learning rate of $1\times10^{-3}$ and a multi-step learning rate decay of $\gamma=0.33$ applied at 100K, 150K, and 180K iterations.
We set the batch size to 512 pixels and optimize the simulated results until they appear to converge, or for 250K iterations for the simulated results and 150K iterations for the captured results. For the weighting of the space carving loss, we use $\lambda_{sc}=10^{-3}$ for the simulated dataset and increase this to $\lambda_{sc}=10^{-2}$ for captured data, which benefits from additional regularization.
We train the network on a single NVIDIA A40 GPU.
Also note that simulated results use RGB histograms, such that $\mathbf{c} \in \mathbb{R}^3$. 
In the captured data $\mathbf{c} \in \mathbb{R}$ because we illuminate the scene with monochromatic laser light.


\section{Multiview Lidar Dataset}

We introduce a first-of-its-kind dataset consisting of simulated and captured multiview data from a single-photon lidar.
A description of the full set of simulated and captured scenes is included in the supplemental, and the dataset and simulation code are publicly available on the project webpage.

\paragraph{Simulated dataset}

\begin{wrapfigure}[19]{r}{2.313 in}
    \vspace{-2em}
    \includegraphics[]{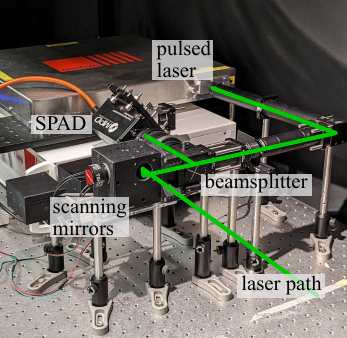}
    \caption{Hardware prototype. A pulsed laser shares a path with a single-pixel SPAD, and the illumination and imaging path are controlled by scanning mirrors.}
    \label{fig:hardware}
\end{wrapfigure}

We create the simulated dataset using a time-resolved version of Mitsuba 2~\cite{royo2022non} which we modify for efficient rendering of lidar measurements.
The dataset consists of one scene from Vicini et al.~\cite{vicini2022differentiable} and four scenes made available by artists on Blendswap (\url{https://blendswap.com/}), which we ported from Blender to our Mitsuba 2 renderer.
The training views are set consistent with the capture setup of our hardware prototype (described below) such that the camera viewpoint is rotated around the scene at a fixed distance and elevation angle, resulting in 8 synthetic lidar scans used for training. 
We evaluate on rendered measurements from six viewpoints sampled from the NeRF Blender test set~\cite{mildenhall2021nerf}.
The renders are used to simulate SPAD measurements by applying the noise model described in Equation~\ref{eq:histogram} and setting the mean number of photon counts to 2850 per occupied pixel and the background counts to 0.001 per bin, which we set to approximate our experimentally captured data.

\paragraph{Hardware prototype.}
To create the captured dataset, we built a hardware prototype (Fig.~\ref{fig:hardware}) consisting of a pulsed laser operating at 532 nm that emits 35 ps pulses of light at a repetition rate of 10 MHz.
The output power of the laser is lowered to $<1$ mW to keep the flux low enough (roughly $150,000$ counts per second on average) to prevent pileup, which is a non-linear effect that distorts the SPAD measurements~\cite{rapp2019dead}. 
The laser shares an optical path with a single-pixel SPAD through a beamsplitter, and a set of 2D scanning mirrors is used to raster scan the scene at a resolution of 512$\times$512 scanpoints.
A time-correlated single-photon counter is used to record the photon timestamps with a total system resolution of approximately 70 ps.

\paragraph{Captured dataset.}
We capture multiview lidar scans of six scenes by placing objects on a rotation stage in front of the scanning single-photon lidar and capturing 20 different views in increments of 18 degrees of rotation. 
For each lidar scan we accumulate photons during a 20 minute exposure time to minimize noise in the transient measurements. 
We bin the photon counts into histograms with 1500 bins and bin widths of 8 ps (all raw timestamp data will also be made available with the dataset).
We set aside 10 views sampled in 36 degree increments for testing and we use 8 of the remaining views for training.
Prior to input into the network for training, we normalize the measurement values by the maximum photon count observed across all views.

\paragraph{Calibration.}
We calibrate the camera intrinsics of the system using a raxel model~\cite{grossberg2005raxel} with corners detected from two scans of checkerboard translated in a direction parallel to the surface normal.
This model calibrates the direction of each ray individually, which is necessary because the 2D scanning mirrors deviate from the standard perspective projection model~\cite{eisert2011mathematical}.
Extrinsics are calibrated by placing a checkerboard on a rotation stage and solving for the axis and center of rotation that best align the 3D positions of the checkerboard corners, where the 3D points are found using the calibrated ray model along with the time of flight from the lidar (see supplemental). 
Overall, accurate calibration is an important and non-trivial task because multiview lidar scans provide two distinct geometric constraints (i.e. stereo disparity and time of flight) that must be consistent for scene reconstruction.

\begin{figure*}[t]
    \centering
    \includegraphics[width=1\textwidth]{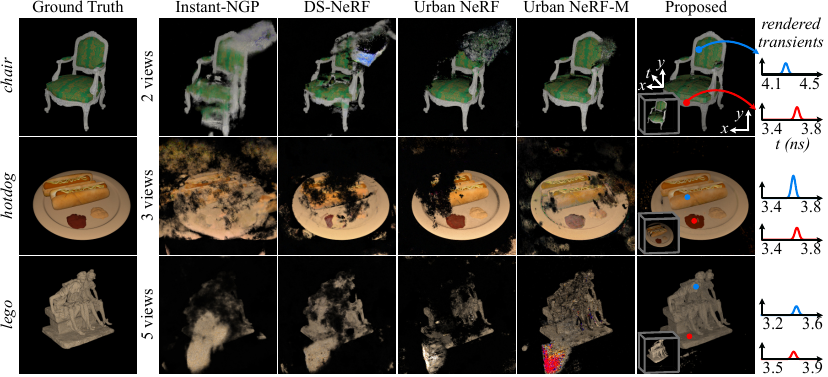}
    \caption{Results on simulated data. We show images from depth-supervised NeRF baselines as well as color images and rendered transients from our method after training on 2, 3, and 5 viewpoints. The proposed method produces cleaner results and generates 3D transients for each viewpoint.}
    \label{fig:simulated}
\end{figure*}

\begin{table*}
    \caption{Simulated results comparing images and depth for the baselines and proposed approach.}
    \label{tab:simulated}
    \centering
    \resizebox{0.9\textwidth}{!}{%
    \begin{tabular}{l|ccc|ccc|ccc}
        \toprule  & \multicolumn{3}{c|}{PSNR (dB)$\,\uparrow$} & \multicolumn{3}{c|}{LPIPS$\,\downarrow$} & \multicolumn{3}{c}{L1 (depth) $\,\downarrow$ }\\ 
        \textbf{Method}                             & 2 views & 3 views & 5 views & 2 views & 3 views & 5 views  & 2 views & 3 views & 5 views \\\midrule
        Instant NGP~\cite{muller2022instant}        & 16.62 & 17.67 & 19.66 & 0.520 & 0.476 & 0.387 & 0.238 & 0.195 & 0.178\\

        DS-NeRF~\cite{deng2022depth}                & \cellcolor{yellow}19.28 & \cellcolor{yellow}19.35 & \cellcolor{yellow}21.07 & \cellcolor{yellow}0.431 &  \cellcolor{yellow}0.436 &  \cellcolor{yellow}0.376 & \cellcolor{yellow}0.109 & \cellcolor{yellow}0.115 & 0.119\\
        Urban NeRF~\cite{rematas2022urban}          & 18.86 & 18.73 & 19.80 & 0.500 & 0.484 & 0.406 & 0.131 & 0.124 & \cellcolor{yellow}0.101\\
        Urban NeRF w/mask~\cite{rematas2022urban}   & \cellcolor{orange}20.91 & \cellcolor{orange}20.81 & \cellcolor{orange}22.34 & \cellcolor{orange}0.410 & \cellcolor{orange}0.382 & \cellcolor{orange}0.339 &  \cellcolor{orange}0.051 &  \cellcolor{orange}0.038 &  \cellcolor{orange}0.029\\
        Proposed                                    & \cellcolor{red}21.38 & \cellcolor{red}23.48 & \cellcolor{red}28.39 & \cellcolor{red}0.172 & \cellcolor{red}0.151 & \cellcolor{red}0.115 & \cellcolor{red}0.015 & \cellcolor{red}0.011 & \cellcolor{red}0.013\\\bottomrule
\end{tabular}}
\end{table*}

\section{Results}
\label{sec:results}

We evaluate our method on the simulated and captured datasets and use transient neural radiance fields to render intensity, depth, and time-resolved lidar measurements from novel views.

\paragraph{Baselines.}
The intensity and depth rendered from our method are compared to four other baseline methods that combine point cloud-based losses with neural radiance fields.
For fairer comparison and to speed up training and inference times, we implement the baselines by incorporating their loss terms into the recently introduced frameworks of NerfAcc~\cite{li2022nerfacc} and Instant-NGP~\cite{muller2022instant} adopted by our method.
We train the following baselines using intensity images (i.e., the photon count histograms integrated over time) along with point clouds obtained from the photon count histograms using a log-matched filter, which is the constrained maximum likelihood depth estimate~\cite{kirmani2014first}.
\begin{itemize}[noitemsep,topsep=0pt,leftmargin=*]
    \item Instant-NGP~\cite{muller2022instant} is used to illustrate performance without additional depth supervision.
    \item Depth-Supervised NeRF (DS-NeRF)~\cite{deng2022depth} incorporates an additional loss term to ensure that the expected ray termination distance in volume rendering aligns with the point cloud points.
    \item Urban NeRF~\cite{rematas2022urban} incorporates the ray-termination loss of DS-NeRF while also adding space carving losses to penalize density along rays before and after the intersection with a point cloud point.
    \item Urban NeRF with masking (Urban NeRF-M) modifies Urban NeRF to incorporate an oracle object mask and extends the space carving loss to unmasked regions, providing stronger geometry regularization (additional details in supplement).
    
\end{itemize}
Prior to input into the network, we normalize the images and apply a gamma correction, which improves network fitting to the high dynamic range data. Finally, after training with each method, we estimate an associated depth map using the expected ray termination depth at each pixel, which is the same metric used in the loss functions of the aforementioned baselines.

\subsection{Simulated Results}
The method is compared to the baselines in simulation across five scenes: \textit{chair}, \textit{ficus}, \textit{lego}, \textit{hot dog}, and \textit{statue}.
In Fig.~\ref{fig:simulated}, we show RGB images rendered from novel views using the baselines and our proposed method after training on two, three, and five views.
More extensive sets of results on all scenes are included in the supplemental.
We find that views rendered from transient neural radiance fields have fewer artifacts and spurious patches of density, as the explicit supervision from the photon count histograms avoids the ill-posedness of the conventional multiview reconstruction problem.

\begin{wrapfigure}[17]{r}{2.6in}
    \vspace{-1em}
    \includegraphics[width=2.6in]{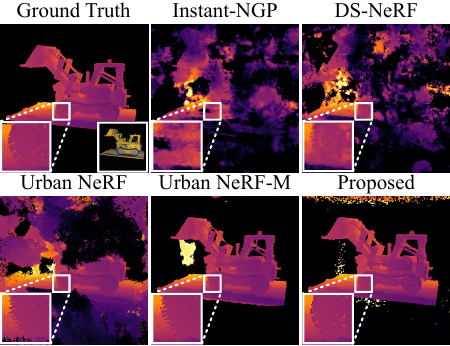}
    \caption{Comparison of depth maps recovered from simulated measurements trained on 5 views of the \textit{lego} scene.} 
    \label{fig:depth}
\end{wrapfigure}

Additional quantitative results are included in Table~\ref{tab:simulated}, averaged across all simulated scenes. 
For the evaluation of rendered RGB images, we normalize and gamma-correct the output of the proposed method and the ground truth in the same fashion as the baseline methods.
Transient NeRF recovers novel views with significantly higher peak signal-to-noise ratio and better performance on the learned perceptual image patch similarity (LPIPS) metric~\cite{zhang2018unreasonable} compared to baselines. Transient measurements provide explicit supervision of the unoccupied spaces in the scene, leading to fewer floating artifacts, and cleaner novel views. 

The depth maps inferred from Transient NeRF are also significantly more accurate than baselines (see Fig.~\ref{fig:depth}).
One key advantage here is that we avoid 
supervision on point clouds obtained by potentially noisy (and thus view-inconsistent) per-pixel estimates of depth.
By training on the raw photon count histograms, the scene's geometry is allowed to converge to the shape that best explains all histograms across all views, resulting in much higher geometric accuracy. 

\begin{figure*}[t]
    \includegraphics[width=1\textwidth]{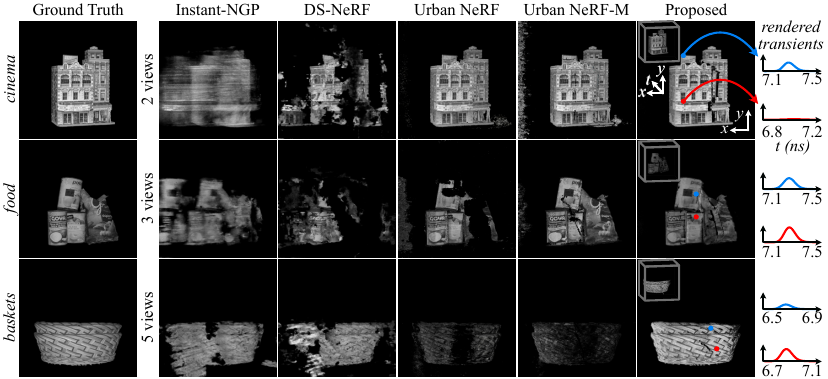}
    \caption{Results on multiview lidar data captured with the hardware prototype and trained with 2, 3, and 5 viewpoints. For the proposed method we show the rendered transients, intensity image, and individual transients for the indicated pixels.}
    \label{fig:captured}
\end{figure*}

\begin{table*}
    \caption{Evaluation of rendered intensity images and depth on captured results.}
    \label{tab:captured}
    \centering
    \resizebox{0.9\textwidth}{!}{%
    \begin{tabular}{l|ccc|ccc|ccc}
        \toprule  & \multicolumn{3}{c|}{PSNR (dB)$\,\uparrow$} & \multicolumn{3}{c|}{LPIPS$\,\downarrow$} & \multicolumn{3}{c}{L1 (depth) $\,\downarrow$} \\ 
        \textbf{Method}                             & 2 views & 3 views & 5 views & 2 views & 3 views & 5 views & 2 views & 3 views & 5 views \\\midrule
        Instant NGP~\cite{muller2022instant}        & \cellcolor{yellow}16.44 & 
        \cellcolor{yellow}16.52 & 
        \cellcolor{yellow}16.39 & 
        \cellcolor{yellow}0.358 & 
        \cellcolor{yellow}0.307 & 0.274
        & 0.115 & 0.076 & 0.053
        \\

        DS-NeRF~\cite{deng2022depth}                & 15.34 & 15.05 & 14.86 & \cellcolor{orange}0.311 &  0.312 &  0.325 & 0.048 & 0.036 & 0.036\\
        Urban NeRF~\cite{rematas2022urban}          & \cellcolor{orange}16.90 & 15.91 & 15.93 & 0.403 & 0.328 &  \cellcolor{yellow}0.231 & \cellcolor{yellow}0.017 & \cellcolor{yellow}0.015 &\cellcolor{yellow} 0.014\\
        Urban NeRF w/mask~\cite{rematas2022urban}   & 15.45 & \cellcolor{orange}18.26 & \cellcolor{orange}19.11 & 0.458 &\cellcolor{orange} 0.269 & \cellcolor{orange}0.191 & \cellcolor{orange}0.014 & \cellcolor{red}0.006 & \cellcolor{red}0.004\\
        Proposed                                    & \cellcolor{red}22.11 & \cellcolor{red}21.83 & \cellcolor{red}22.72 & \cellcolor{red}0.271 & \cellcolor{red}0.212 & \cellcolor{red}0.172 & \cellcolor{red}0.005 & \cellcolor{red}0.006 & 
        \cellcolor{orange}0.010 \\\bottomrule
\end{tabular}}
\end{table*}

\subsection{Captured Results}
In Fig.~\ref{fig:captured} we show rendered novel views of intensity images from our method and baselines trained on captured data with two, three, and five views. 
Results are shown on the \textit{cinema}, \textit{food}, and \textit{baskets} scenes (additional results in the supplemental).
The proposed approach results in fewer artifacts and the rendered intensity images are more faithful to reference intensity images captured from the novel viewpoint.
Quantitative comparisons of our method to baselines on captured data are shown in Table~\ref{tab:captured}; note that we do not have access to ground truth depth for captured data and instead use depth from a log-matched filter on the ground truth transient.
We find that the method outperforms the baselines in terms of PSNR and LPIPS of intensity images rendered from novel views. 
While performance on captured data does not improve as much as observed on simulated data with increasing numbers of viewpoints, we attribute this to small imperfections ($\approx$1 mm) in the alignment of the lidar scans after estimating the camera extrinsics.

\begin{wrapfigure}[14]{r}{3in}
    \vspace{-1.5em}
    \includegraphics[width=3in]{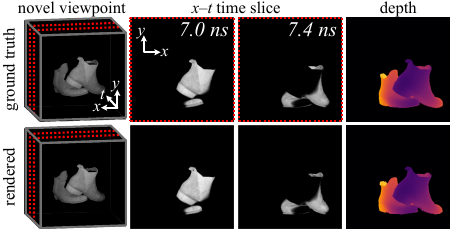}
\caption{Comparison between reference and novel views of lidar measurements, intensity slices, and depth. The $x$--$y$ intensity slices are visualized for times indicated by the red dashed lines.}
    \label{fig:captured_transients}
\end{wrapfigure}

Since DS-NeRF trains explicitly on depth without additional regularization, it is especially sensitive to camera perturbations and can be outperformed in some cases by Instant NGP which has no additional geometry constraints.
Our approach appears somewhat less sensitive to these issues, perhaps because geometry regularization is done implicitly through a photometric loss on the lidar measurements. 

We notice some degradation in depth accuracy relative to simulation, likely due to imperfections in the estimated extrinsics.
Sub-mm registration of the lidar measurements would likely improve results, but achieving such precise registration is non-trivial and beyond the scope of our current work.
Finally, in Fig.~\ref{fig:captured_transients} we compare captured measurements to rendered transients and depth rendered for the \textit{boots} scene trained on 2 viewpoints. 
We recover the time-resolved light propagation from a novel view, shown in $x$--$y$ slices of the rendered transients over time.
The depth map recovered from the novel view appears qualitatively similar to the ground truth (estimated from captured measurements using a log-matched filter~\cite{rapp2017few}).
We show additional 3D reconstruction results in the supplemental. 

\section{Discussion}
Our work brings NeRF to a new dimension of imaging at transient timescales, offering new opportunities for view synthesis and 3D reconstruction from multiview lidar.
While our work is limited to modeling the direct reflection of laser light to perform lidar view synthesis, our dataset captures much richer light transport effects, including multiple bounces of light and surface reflectance properties that could open avenues for future work. 
In particular, the method and dataset may help enable techniques for intra-scene non-line-of-sight imaging~\cite{velten2012recovering,o2018confocal,lindell2019wave,liu2019non} (i.e., recovering geometry around occluders within a scene), and recovery of the bidirectional reflectance distribution function via probing with lidar measurements~\cite{naik2011single}.

Our method is also limited in that we do not explore more view synthesis in more general single-photon imaging setups, such as when the lidar and SPAD are not coaxial; we hope to explore these configurations in future work.
The proposed framework and the ability to render transient measurements from novel views may be especially relevant for realistic simulation for autonomous vehicle navigation, multiview remote sensing, and view synthesis of more general transient phenomena.

\begin{ack}
Kiriakos N. Kutulakos acknowledges the support of the Natural Sciences and Engineering Council of Canada (NSERC) under the RGPIN and RTI programs.
David B. Lindell acknowledges the support of the NSERC RGPIN program.
The authors also acknowledge Gordon Wetzstein and the Stanford Computational Imaging Lab for loaning the single-photon lidar equipment.
\end{ack}

{\small
\bibliographystyle{unsrtnat}
\bibliography{refs}
}



\section*{Supplementary material}

# Transient Neural Radiance Fields for Lidar View Synthesis and 3D Reconstruction
# –Supplementary Material–

**Anagh Malik**[1,2]
anagh@cs.toronto.edu

**Parsa Mirdehghan**[1,2]
parsa@cs.toronto.edu

**Sotiris Nousias**[1]
sotiris@cs.toronto.edu

**Kiriakos N. Kutulakos**[1,2]
kyros@cs.toronto.edu

**David B. Lindell**[1,2]
lindell@cs.toronto.edu

[1]University of Toronto      [2]Vector Institute

## 1  Hardware Prototype

To create the captured dataset, we built a hardware prototype consisting of a pulsed laser (NKT Photonics Katana 05HP) operating at 532 nm that emits 35 ps pulses of light at a repetition rate of 10 MHz. The output power of the laser is lowered to $< 1$ mW to keep the flux low enough (roughly $150{,}000$ counts per second on average) to prevent pileup, which is a non-linear effect that distorts the SPAD measurements [1]. The laser emits polarized light that passes through a polarizing beam splitter (Thorlabs PBS251), to a set of 2D scanning mirrors (Thorlabs GVS012). The mirrors are controlled by a multifunction I/O device (NI-DAQ USB-6343) and are used to scan scenes at a spatial resolution of $512{\times}512$ at a rate of 0.1 frames per second. The laser shares an optical path through the beam splitter with a single pixel SPAD (Micro Photon Devices PDM series SPAD) with a $50~\mu$m $\times$ $50~\mu$m active pixel area. Photons detected by the SPAD are correlated with a sync signal from the laser using a time-correlated single photon counter (TCSPC) to measure the photon arrival timestamps.

We place the scanned scenes on a rotation stage (Parker Motion/Parker 6K4 Compumotor) in front of the scanning single-photon lidar, allowing us to capture different viewpoints by rotating the scene.

Each scan consists of 20 minutes of total exposure time, but we save out all collected photon timestamps individually so that any desired exposure time can be emulated by accumulating photons over the desired time window in post processing (i.e., for future applications of the dataset). We set the bin width of the photon count histograms to 8 ps and the number of bins is 1500. The entire data acquisition is controlled using custom-developed MATLAB software on a desktop PC, and we capture six scenes with varying geometry, texture, and material properties. For each scene, we capture views in 18 degree increments of the rotation stage, resulting in a 360 degree capture. We set aside 8 views for training, and 10 separate views sampled in 36 degree increments comprise the test split. Prior to input into the network for training, we normalize the measurement values by the maximum photon count observed across all views.

**Ambient illumination.**   Our hardware setup is relatively robust to ambient illumination because we place a laser-line spectral filter (Thorlabs FL532-10) in front of the SPAD, which attenuates ambient illumination by a factor of 10,000. Under indoor lighting, we observe roughly 300-3,000 photon counts per second, depending on the albedo of the target, which is $< 2\%$ of the detected laser photons (150,000 counts per second). However, operation in much brighter environments (e.g., outdoors under direct sunlight) would result in non-negligible background counts. In that case, it is possible to use

other laser wavelengths (e.g., 1550 nm ), where sunlight is heavily attenuated because of absorption by the atmosphere.

## 2 Calibration of the Hardware Prototype

Here we describe the method used to find the extrinsics and intrinsics defining the captured dataset. An overview of the method can be found in the Algorithm 1.

**Intrinsics**. We calibrate the camera intrinsics of the system using the raxel model [2], which maps each pixel to a 3D ray direction. To find the ray directions, we place a checkerboard on a translation stage and use the lidar system to capture two images of this checkerboard before and after translating 17 cm by in the direction of the surface normal of the board (this was the maximum distance permitted by the translation stage and our optical table layout).

Checkerboard corners are detected in the two images using OpenCV's *findChessboardCorners* [3] with subpixel refinement. In order to improve robustness to distortion, we further refine the corner detection by fitting a second-order polynomial to corners along vertical lines (which we found to be a good fit to model the observed distortion), and we use a first-order fit to horizontal lines, which we found to fit the data well. We set the corner positions to the intersections of the set of fitted lines.

The pixel coordinates of the detected checkerboard corners are then used to define a 3D coordinate system. We set the origin of the coordinate system to the upper-left corner of the nearest checkerboard, and coordinates on the far checkerboard are calculated based on the known translation in $z$, with $x$ and $y$ coordinates given using the size of the checkerboards (4.2 mm). We associate ray directions to each pixel by finding the points of intersection of the ray with the two checkerboards. Specifically, we linearly interpolate the mapping from pixel values to 3D coordinates at each checkerboard corner to retrieve a dense mapping from pixels to 3D coordinates on each board. Then, for a given pixel the ray direction is given by a simple subtraction of the 3D intersection points on each board.

**Extrinsics**. Our captured dataset consists of photon count histograms from six different scenes, each with 20 different views. We use a rotation stage to move the scenes one full revolution in 18 degree increments. The resulting views are equivalent to capturing a stationary scene with cameras that are rotated about the center of rotation of the stage. Since all scenes are captured identically in this fashion, we determine the camera extrinsics once for all scenes.

We also calibrate for an additional offset parameter to finetune the time-of-flight-delays recorded by the lidar system. This accounts for the unknown time delay between the time that the time-correlated single-photon counter receives the sync signal from the laser and the time that the laser pulse reaches the center of projection of the scanning mirrors. In other words, this offset accounts for when "time zero" should occur in the photon count histograms; we roughly calibrate for this value by placing a target directly in front of the galvo mirrors, and we fine tune this offset via optimization as detailed below.

Given that each view is captured using a controlled rotation, the extrinsics can be determined by identifying the 3D axis of rotation and 3D center of rotation. To estimate these parameters, we use a two step procedure. First, we capture lidar scans of a checkerboard placed on the rotation stage and rotated to 6 different positions in 9 degree increments. We convert the lidar scans to a point cloud using the raxel model and lidar time of flight, and then a coarse solution is obtained by fitting planes to the point clouds and finding the center and axis of rotation that align the plane normals. Second, we detect corners of the checkerboards and find the corresponding 3D points. Then, we optimize for the center of rotation, axis of rotation, and the 1D offset parameter that best align the 3D checkerboard corners.

Specifically, we implement a routine in PyTorch [4] to align the 3D checkerboard corners using the Rodrigues formula given below.

$$\mathbf{v}' = (\mathbf{v} - \mathbf{c}) \cos \theta + (\mathbf{a} \cdot (\mathbf{v} - \mathbf{c}))(1 - \cos \theta) \mathbf{a} + (\mathbf{a} \times (\mathbf{v} - \mathbf{c})) \sin \theta + \mathbf{c}. \tag{S1}$$

Here, $\mathbf{v}$ is the 3D point to be rotated, $\mathbf{c}$ is the center of rotation, $\mathbf{a}$ is the axis of rotation, and $\mathbf{v}'$ is the rotated point. We minimize the objective function given as

$$\mathcal{L}_{\text{align}} = \sum_p \sum_{[i,j] \in \mathcal{C}} \|\mathbf{v}'^{(i)}_p - \mathbf{v}'^{(j)}_p\|_2^2, \tag{S2}$$

where $p$ indexes the checkerboard corner points and $[i, j]$ index all pairs of checkerboards. Thus we penalize the distance between corresponding points between all checkerboards. The optimization is performed using LBFGS [5]. After optimization, we use the resulting center point and rotation matrices (i.e., by rotating around **a** by increments of 18 degrees) to define the camera extrinsics for each view.

---

**Algorithm 1:** Calibration overview.

**Intrinsics Calibration**
**Data:** Two scans of a checkerboard before and after translation.

1. Perform sub-pixel detection of checkerboard corners using OpenCV and polynomial line fitting.
2. Use the detected checkerboard corners to initialize a 3D coordinate system.
3. Compute per-pixel 3D coordinates for each checkerboard scan by interpolating the mapping from corner pixel coordinates to 3D coordinates.
4. Compute the ray directions by subtracting the per-pixel 3D coordinates obtained for the checkerboard scan before and after translation.

**Extrinsics Calibration**
**Data:** Six scans of a checkerboard rotated using the rotation stage.

1. Convert lidar scans to point clouds using the time of flight and intrinsics.
2. Fit planes and determine initial center and axis of rotation to align the surface normals.
3. Detect 3D points corresponding to checkerboard corners.
4. Refine the initial center and axis of rotation via Equations S1 and S2.

---

## 3 Additional Implementation Details

**Network Architecture**. In this section we describe the main network architecture. We extend the NerfAcc [6] framework and the provided implementation of Instant-NGP (INGP) [7]. All network hyperparameters are shared between our proposed method and the baselines unless otherwise stated. We set the number of hash feature grids for INGP to 16 and set the feature size to 2. The resolution of the coarsest grid is set to 16 and each subsequent grid has 2 times finer resolution. The base MLP has width 64 with 1 hidden layer to map to density and a latent vector. Another MLP with 2 hidden layers and 64 hidden units maps from the latent vector and viewing direction to the view-dependent radiance.

We use the occupancy grid from NerfAcc with a resolution of $128^3$. The occupancy grid is employed to remove samples along the ray based on their density for the sake of efficiency. The grid is binarized using a occupancy value of $10^{-3}$.

For the captured data, we find that setting the binarization threshold to $10^{-5}$ for the first 3,000 iterations before reverting back to the standard $10^{-3}$ value helps to avoid an overly aggressive removal of density that erodes the surface of reconstructed objects. In addition, we incorporate pruning based on transmittance, removing any samples that register a transmittance of 0 to speed up rendering. Finally, all rays are rendered by sampling 4,096 points along each ray, and these points are pruned according to their occupancy and transmittance values, as previously discussed.

We set the bounding box used in INGP as follows for the simulated and captured data. For the simulated data the bounding box extents are set to -1.5 to 1.5 across all dimensions and methods. For captured data, we use a -0.4 to 0.4 bounding box for Transient NeRF; we slightly shrink the bounding box for baselines run on captured data to -0.3 to 0.3, which we find helps remove some spurious regions of density and improves the baseline results.

**Rendering Equation** Given samples $t_1, t_2, \cdots t_N$ along a camera ray, we render a histogram bin $\tau[n]$ for the ray as

$$\tau[n] = \sum_{i \mid \frac{t_i + t_{i+1}}{2} \in \mathcal{T}_n} T_i^2 \left(1 - \exp\left(-\sigma_i \delta_i\right)\right) \frac{\mathbf{c}_i}{(\frac{1}{2}(t_i + t_{i+1}))^2}, \text{ where } T_i = \exp\left(-\sum_{j=1}^{i-1} \sigma_j \delta_j\right), \tag{S3}$$

where $\sigma_i, \mathbf{c}_i$ are the density and radiance outputs of the network, $\delta_i = t_{i+1} - t_i$ is the distance between consecutive samples and $\mathcal{T}_n$ is the resolution of the histogram bin, in units of distance. The sum is taken over all consecutive samples whose midpoints fall within $\mathcal{T}_n$.

**Optimization settings and run time**. The time required for training is strongly influenced by the number of samples used for the spatial filter, as discussed in the main text. Specifically, each image pixel is rendered by computing a weighted integral over the radiance for a particular region of space. We compute this integral during training by stochastically sampling one or more ray directions per pixel with probability determined by the weighting and support of the spatial filter [8]. For the simulated dataset, we initially sample a single ray for the first 2000 iterations, then subsequently double this number every 2000 iterations until we reach a maximum of 30 rays sampled per pixel. For the captured dataset we use a single ray sample per pixel per iteration; we find that increasing the number ray samples results in longer training times without significantly improved performance.

Training takes roughly eight hours to converge for the simulated dataset (250K iterations) and two hours to converge for captured data (150K steps).

**Depth calculation**. We follow the NeRF convention in calculating depth for all baseline methods as

$$d = \sum_i \sigma(t_i) T(t_i) \frac{t_i + t_{i+1}}{2}, \tag{S4}$$

where $t_i$ is the distance from the ray origin to a sample along the ray, $T_i$ is the transmittance, and $\sigma$ is the density. Intuitively, this equation calculates the expected ray termination distance.

Since our method incorporates a spatial filter to more accurately model the footprint of the illumination spot, a single ray can result in measurements from multiple surfaces (e.g., if a pixel integrates over a depth discontinuity). To avoid multiple depths in the measurement from skewing the estimate of the expected ray termination distance, we use the following equation to compute depth.

$$d = \arg\max_{t_i} \sigma(t_i) T(t_i). \tag{S5}$$

Thus, we find the depth at which the maximum probability of ray termination occurs.

We note that in the conventional NeRF depth rendering formula of Equation S4, regions of low density become less visible in most visualizations because their depth tends towards zero (i.e., it is weighted by the density, which is close to zero, but typically not uniformly zero for empty regions). However, in Equation S5, no such weighting exists, resulting in visualizations that appear noisier because depths for regions with low (but non-zero) density are not automatically suppressed. Thus for visualization of the depth maps we weight the depth as follows.

$$d = \left(\sum_i \sigma(t_i) T(t_i)\right) \arg\max_{t_i} \sigma(t_i) T(t_i). \tag{S6}$$

We use this equation to visualize depth maps, as we find they are more comparable to those rendered with the conventional NeRF formulation (though the density and transmittance weighting results in them being slightly less accurate).

### 3.1 Baseline Implementation Details

Four different baselines are implemented to evaluate our proposed method. To isolate the impact of different loss functions and speed up the training, we implemented the specific loss functions from each of the following methods in the framework of NerfAcc with the Instant-NGP backbone.

For all baselines, we generate the ground truth intensity images by first integrating the transients over the time dimension and normalization by a scale factor to shift the image values to lie close to within

[0, 1]. We apply gamma correction to tonemap the resulting high dynamic range intensity images prior to training.

**Instant-NGP [7].** The Instant-NGP model is trained with photometric loss defined as the total squared error between the rendered intensities and the pixel colors from the input images:

$$\mathcal{L}_{\text{photo}} = \sum_{\mathbf{r} \in \mathcal{R}} \|\widetilde{\mathbf{C}}(\mathbf{r}) - \mathbf{C}(\mathbf{r})\|_2^2, \tag{S7}$$

where $\widetilde{\mathbf{C}}$ and $\mathbf{C}$ denote the ground-truth pixel color and the predicted value, respectively. $\mathcal{R}$ specifies the set of *active* rays that have non-zero opacity values; this set of rays is updated as training progresses to accelerate optimization by pruning rays that do not contribute to the rendering [6].

**Depth-Supervised NeRF [9].** Following the depth supervision idea proposed in [9], we train a new model that incorporates depth error loss in addition to the photometric loss above:

$$\mathcal{L}_{\text{depth}} = \frac{1}{|\mathcal{R}'|} \sum_{\mathbf{r} \in \mathcal{R}'} \left(\widetilde{d}(\mathbf{r}) - d(\mathbf{r})\right)^2. \tag{S8}$$

Here, $\widetilde{d}(\mathbf{r})$ denotes the lidar depth for ray $\mathbf{r}$ obtained from the captured transients using a log-matched filter [10], $d(\mathbf{r})$ is the predicted depth value computed from Equation S4, and $\mathcal{R}'$ specifies the set of rays that intersect with the object.

The final training loss is defined as $\mathcal{L} = \mathcal{L}_{\text{photo}} + \lambda_{\text{depth}} \mathcal{L}_{\text{depth}}$. In the simulated and captured results, $\lambda_{\text{depth}}$ is set to $0.005$ and $0.0075$, respectively.

**Urban NeRF [11].** We further incorporated the *line-of-sight lidar loss* $\mathcal{L}_{\text{sight}}$, proposed in [11], to encourage the densities to be concentrated near the lidar points. This loss comprises two terms. The first term penalizes any density between the ray origin and the lidar point:

$$\mathcal{L}_{\text{empty}} = \frac{1}{|\mathcal{R}'|} \sum_{\mathbf{r} \in \mathcal{R}'} \left[ \int_{t_n}^{\widetilde{d}-\epsilon} w(t)^2 dt \right]. \tag{S9}$$

In this equation, $\mathcal{R}'$ is the set of rays that intersect with the object; $w(t)$ is the volume rendering integration weights defined as $\sigma(t)T(t)$; $t_n$ denotes the near bound of the ray; and $\epsilon$ specifies a neighbourhood around the surface point.

The second term, on the other hand, encourages the model to increase the densities in the bounded region around the surface point:

$$\mathcal{L}_{\text{near}} = \frac{1}{|\mathcal{R}'|} \sum_{\mathbf{r} \in \mathcal{R}'} \left[ \int_{\widetilde{d}-\epsilon}^{\widetilde{d}+\epsilon} (w(t) - \mathcal{K}_\epsilon(t - \widetilde{d}))^2 dt \right], \tag{S10}$$

where $\mathcal{K}_\epsilon$ is a truncated Gaussian defined as $\mathcal{N}(0, (\epsilon/3)^2)$ [12].

The final training loss for this model is defined as:

$$\mathcal{L} = \mathcal{L}_{\text{photo}} + \lambda_{\text{depth}} \mathcal{L}_{\text{depth}} + \lambda_{\text{sight}} \mathcal{L}_{\text{sight}}, \tag{S11}$$

where $\mathcal{L}_{\text{sight}} = \mathcal{L}_{\text{empty}} + \mathcal{L}_{\text{near}}$. The parameters $(\lambda_{\text{depth}}, \lambda_{\text{sight}})$ are set to $(0.0001, 0.005)$ for both the simulations and captured experiments.

Following the instructions from the original paper, we applied exponential decay to the value of $\epsilon$ throughout the optimization, which encourages the density of the radiance field to fall within a progressively smaller support, improving convergence. We initialize $\epsilon$ to a value $\epsilon_{\max}$, and every 7000 steps we multiply by a factor of $0.8$, until it decays to $\epsilon_{\min}$. Parameters $(\epsilon_{\max}, \epsilon_{\min})$ are set to $(1.5, 0.025)$ and $(4.0, 0.05)$ for the captured and simulated results, respectively.

**Urban NeRF with Masking.** The depth-related loss terms $\mathcal{L}_{\text{depth}}$ and $\mathcal{L}_{\text{sight}}$ used in the previous two models are evaluated only for the rays that intersect with the object; however, we find that the included regularization terms do not prevent spurious regions of density that appear for pixels that fall outside the support of the object on the image plane.

We attempt to improve the baseline results further by using modified versions of $\mathcal{L}_{\text{photo}}$ and $\mathcal{L}_{\text{sight}}$ which sum over active rays $\mathcal{R}$. Specifically we use an oracle object mask (i.e., a ground truth mask

segmenting the object in each training view) and set the ground truth depth of all background pixels to zero. This loss encourages the density to be uniformly zero along background rays and helps to remove spurious clouds of density that tend to materialize in the other baseline results during training. We note that this masking loss is directly analogous to the *sky modeling* loss proposed in Urban-NeRF [11], which penalizes density along rays that intersect with the sky as determined using a semantic segmentation network.

For both the simulated and captured results, we hand-annotate the oracle object masks by carefully thresholding the intensity images and applying morphology operations to fill in holes in the mask.

## 4 Supplemental Results

### 4.1 Dataset

**Simulated dataset.** To create the simulated multiview lidar dataset, we modify the non-line-of-sight Mitsuba 2 rendering codes of Royo et al. [13]. While the original codebase enables one to simulate the effect of illuminating a single point in the scene with a laser and imaging other points, our approach requires rendering a frame where each pixel images an area of the scene that is illuminated with a coaxial, collimated light source. To this end, introduce a new light source as well as additional rendering options such that each pixel is rendered independently with its own coaxial light source. We also incorporate a Gaussian reconstruction filter [8] along the time dimension to avoid aliasing or stair-stepping artifacts in regions with fine variations in depth.

We choose the variance of the Gaussian temporal filter and spatial filters to produce transients that measure a laser pulse with similar temporal and spatial support as the captured results. Specifically, we set the variance of the temporal and spatial filters to $3$ and $0.15$ respectively. We simulate photon count histograms 1200 bins and each bin has a width of approximately $\sim 30 ps$.

We subsample 8 training viewpoints from the simulated lidar measurements rendered at 36 degree increments along a circle centered at the origin. Specifically, we sample 2 views separated by 180 degrees; 3 views separated by 90 and 180 degrees (i.e., a superset of the 2 views), and 5 views uniformly separated by 72 degrees. Note that the 5 views are not a superset of the viewpoints used when training on 2 or 3 lidar measurements. For testing, we use 6 viewpoints from the NeRF Blender test set that surround the object.

**Captured dataset.** The captured dataset consists of 20 multiview single-photon lidar scans of six scenes. The scenes consist of everyday objects and figurines (see Fig. S1). We record photon timestamps at 4 ps resolution for each scene, with measurements of each view being captured during an exposure period of 20 minutes. While we use all photon timestamps in the photon count histograms used for the proposed method, access to the photon timestamps also allows synthesizing histograms with arbitrary exposure time, which will make the dataset useful for a wide array of follow-on work.

We use raw photon count histograms with 4096 bins and bin widths of 4 ps. To decrease the memory required for training, we crop and downsample the histograms to 1500 bins with 8 ps resolution.

Measurements are captured in the low-flux regime to avoid non-linear distortions due to pile-up. A detailed breakdown of the photon counts and acquisition parameters for each captured scene is provided in Table S1.

Table S1: Photon counts per scene for the captured dataset.

| scene name | spatial resolution | histogram bins | exposure time/view | avg. counts/view | avg. counts/sec |
|---|---|---|---|---|---|
| *baskets* | 512× 512 | 1500 | 20 min | $9.04 \times 10^7$ | $7.53 \times 10^4$ |
| *boots* |  |  |  | $7.39 \times 10^7$ | $6.16 \times 10^4$ |
| *carving* |  |  |  | $5.60 \times 10^7$ | $4.66 \times 10^4$ |
| *chef* |  |  |  | $2.56 \times 10^8$ | $2.13 \times 10^5$ |
| *cinema* |  |  |  | $1.51 \times 10^8$ | $1.26 \times 10^5$ |
| *food* |  |  |  | $1.44 \times 10^8$ | $1.20 \times 10^5$ |

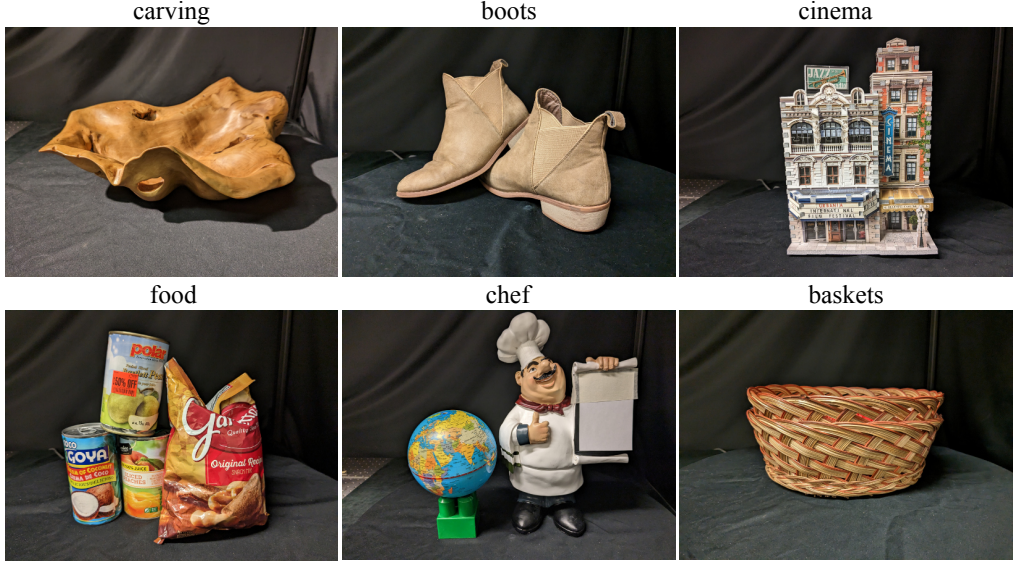

Figure S1: The captured dataset. The captured datset consists of 20 multiview single-photon lidar scans of six scenes.

## 4.2 Depth Evaluation

DS-NeRF and Urban NeRF calculate depth differentiably using the integral along the ray (i.e., the expected ray termination distance) and incorporate this in their loss functions. Since the baseline methods are supervised with this depth estimate, we opted to use the same approach for evaluation.

For completeness, we provide an updated version of the L1 depth in Table S2 calculated for all methods using the maximum ray termination probability as described in Equation S5. Baseline performance using this method of depth estimation is indeed improved for most of the methods, though Transient NeRF still performs best.

Table S2: Depth evaluation using maximum probability of ray termination for simulated and captured scenes.

| | Simulated ↓ | | | Captured ↓ | | |
|---|---|---|---|---|---|---|
| **Method** | 2 views | 3 views | 5 views | 2 views | 3 views | 5 views |
| Instant NGP [30] | 0.151 | 0.202* | 0.184* | 0.021 | 0.021 | 0.024 |
| DS-NeRF [26] | 0.102 | 0.124* | 0.136* | 0.024 | 0.026 | 0.033 |
| Urban NeRF [4] | 0.128 | 0.124 | 0.100 | 0.017 | 0.016* | 0.013 |
| Urban NeRF w/mask† | 0.039 | 0.038 | 0.020 | 0.014 | 0.006 | 0.004 |
| Proposed | 0.015 | 0.011 | 0.013 | 0.006 | 0.006 | 0.010 |

\* Using maximum probability of ray termination is worse than the expected termination distance (Equation S4).
† Requires ground truth segmentation mask.

## 4.3 Intermediate Rendering Results

In Fig. S2, we show intermediate results from the rendering pipeline: the density from the network and the raw radiance output. We plot all these quantities versus the histogram bin number, which we find by simply discretizing the distances of the samples along the rays and allocating the quantities to the appropriate bin.

The rendered densities are very dissimilar from the rendered transients, partly due to the transients being effected by the transmittance, but also because these values are visualized before applying the temporal filter.

Note these plots show renders for a single ray, and we are not integrating over multiple rays to account for the pixel footprint. Also, we plot the normalized density and radiance along the ray since these quantities vary significantly across different pixels in the image.

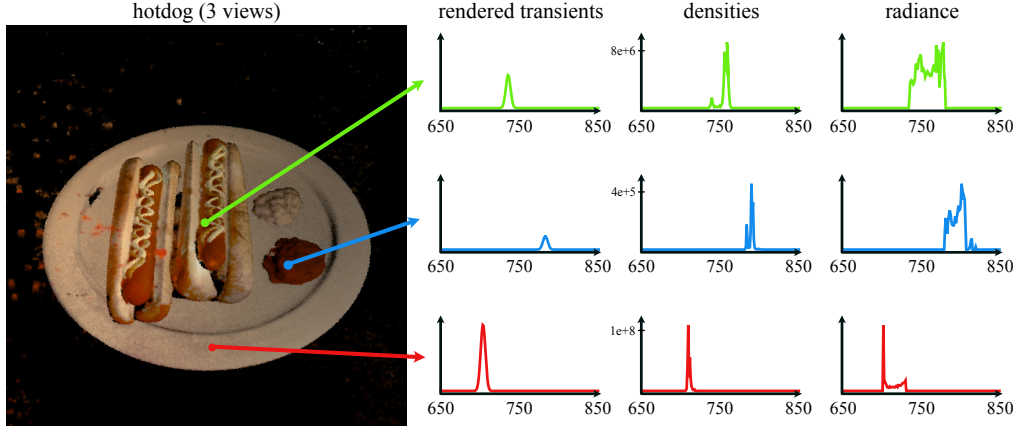

Figure S2: Rendered transients, densities, and radiance plotted versus bin number for rays represented in the rendered image of the *hotdog* scene trained on three views. We normalize the densities and radiance values for visualiation and plot the unnormalized transients.

### 4.4 Ablation Studies

**Space carving, temporal filter and normals.** In Table S3 we show ablation studies of our method calculated on the captured *cinema* scene. Specifically, we include quantitative results of our method without the space carving loss (w/o SC), without accounting for the laser profile (w/o TF) and the proposed method while accounting for normals (w/ normals).

The space carving loss appears to especially improve performance in the five view case. It helps eliminate spurious clouds of density, which improves reconstruction quality for held-out test views.

We also note the importance of explicitly accounting for the laser pulse width and system jitter. This is especially noticeable for the two view results, where the depth becomes highly skewed without this component. We attribute this effect to a thickening of the density representing the object surface; the optimization appears to converge to this solution to explain the temporal spread of the returning light captured in the photon count histograms. Properly accounting for the laser pulse and system jitter essentially deconvolves the temporal response of the lidar system, resulting in thin sheets of density that accurately localize the object surface.

We also ablate using the normals by adding cosine factors to the rendering equation (Eq. 3) for the Cinema scene. Estimating the normals requires a noisy finite-difference operation on the predicted density values, which we found results in speckle-like artifacts in the novel views. Instead, we model the effects of incidence angle by conditioning the neural representation on view direction.

Table S3: Ablation studies on the proposed method for the captured cinema scene. We present results while omitting any temporal filtering (w/o TF), omitting the space carving loss (w/o SC) and with modelling normals using cosine factors (w/ normals).

| Method | PSNR (dB) ↑ | | | LPIPS ↓ | | | SSIM ↑ | | | L1 (depth) ↓ | | |
| --- | --- | --- | --- | --- | --- | --- | --- | --- | --- | --- | --- | --- |
| | 2 views | 3 views | 5 views | 2 views | 3 views | 5 views | 2 views | 3 views | 5 views | 2 views | 3 views | 5 views |
| Proposed w/o TF | 16.55 | 20.85 | 20.02 | 0.346 | 0.225 | 0.209 | 0.589 | 0.837 | 0.823 | 0.022 | 0.007 | 0.009 |
| Proposed w/o SC | 20.63 | 21.09 | 20.12 | 0.207 | 0.200 | 0.217 | 0.855 | 0.855 | 0.840 | 0.007 | 0.009 | 0.020 |
| Proposed w/ normals | 20.88 | 20.41 | 24.62 | 0.332 | 0.268 | 0.216 | 0.825 | 0.815 | 0.867 | 0.006 | 0.007 | 0.007 |
| Proposed | 21.61 | 21.66 | 25.12 | 0.281 | 0.245 | 0.178 | 0.850 | 0.812 | 0.879 | 0.006 | 0.006 | 0.007 |

**Pixel footprint.** We ablate modelling the pixel footprint (see Table S4). When the laser spot and sensor footprint pass over a depth discontinuity, we observe two peaks in the resulting photon count histogram (corresponding to the two depths across the discontinuity). Assuming an ideal spot (i.e., using a single ray) completely fails to model this effect. Training and rendering the *lego* scene (which has many depth discontinuities) with the ideal spot model results in much worse performance across novel views.

8

Table S4: Simulated results comparing the proposed approach on the *Lego* scene with and without modeling the laser spatial footprint (w/o SF).

| Method | PSNR (dB) ↑ | | | LPIPS ↓ | | | L1 (depth) ↓ | | |
|---|---|---|---|---|---|---|---|---|---|
| | 2 views | 3 views | 5 views | 2 views | 3 views | 5 views | 2 views | 3 views | 5 views |
| Proposed w/o SF | 20.01 | 23.39 | 25.08 | 0.195 | 0.147 | 0.170 | 0.119 | 0.065 | 0.041 |
| Proposed | 20.64 | 23.63 | 25.18 | 0.190 | 0.161 | 0.192 | 0.023 | 0.011 | 0.008 |

**HDR-informed loss function.** We provide an ablation study of the HDR-informed loss function in Table S5 on the *chef* and *food* scenes. Incorporating this loss provides some improvement by preventing very bright regions from dominating the loss. Moreover, in the case of 5 input views, performance without the HDR-informed loss drops due to specular highlights appearing in one view, but not an overlapping nearby training view (e.g., on the globe in the *chef* scene or the bag of chips in the *food* scene). Without the HDR-informed loss, the network has difficulty modeling the large variation in radiance between views, and so the optimization produces spurious patches of density to model this view-dependent effect (despite regularization with the space carving loss).

Table S5: Ablation study of the HDR-informed loss function for the captured *Chef* and *Food* scenes.

| Scene/Method | PSNR (dB) ↑ | | | LPIPS ↓ | | | SSIM ↑ | | | L1 (depth) ↓ | | |
|---|---|---|---|---|---|---|---|---|---|---|---|---|
| | 2 views | 3 views | 5 views | 2 views | 3 views | 5 views | 2 views | 3 views | 5 views | 2 views | 3 views | 5 views |
| *Chef* w/o HDR | 18.19 | 18.09 | 13.54 | 0.360 | 0.324 | 0.345 | 0.767 | 0.773 | 0.643 | 0.004 | 0.005 | 0.025 |
| *Chef* w/ HDR | 19.27 | 18.14 | 19.55 | 0.334 | 0.338 | 0.276 | 0.775 | 0.747 | 0.811 | 0.006 | 0.007 | 0.013 |
| *Food* w/o HDR | 23.08 | 23.98 | 16.93 | 0.305 | 0.184 | 0.263 | 0.827 | 0.881 | 0.722 | 0.007 | 0.006 | 0.032 |
| *Food* w/ HDR | 23.40 | 23.78 | 22.09 | 0.286 | 0.205 | 0.154 | 0.826 | 0.879 | 0.873 | 0.006 | 0.007 | 0.013 |

## 4.5 Simulated Results

The paper focuses on results for sparse views (i.e., 2, 3, and 5 views) since this is the regime where depth supervision improves the most over only using 2D supervision. However in Table S6 we show results of the proposed method and the baselines trained on 10 simulated training views sampled at equal angles around the Lego scene, and our approach still outperforms the baselines when evaluated on the same test views.

Table S6: Simulated results for 2, 3, 5, and 10 training views on the *Lego* scene.

| Method | PSNR (dB) ↑ | | | | LPIPS ↓ | | | | L1 (depth) ↓ | | | |
|---|---|---|---|---|---|---|---|---|---|---|---|---|
| | 2 views | 3 views | 5 views | 10 views | 2 views | 3 views | 5 views | 10 views | 2 views | 3 views | 5 views | 10 views |
| Instant NGP [30] | 16.04 | 17.46 | 16.46 | 21.73 | 0.591 | 0.544 | 0.479 | 0.216 | 0.224 | 0.236 | 0.195 | 0.101 |
| DS-NeRF [26] | 17.66 | 19.45 | 17.21 | 22.83 | 0.500 | 0.476 | 0.482 | 0.302 | 0.136 | 0.127 | 0.193 | 0.068 |
| Urban NeRF [4] | 19.59 | 19.75 | 16.19 | 21.36 | 0.519 | 0.522 | 0.503 | 0.368 | 0.105 | 0.100 | 0.151 | 0.074 |
| Urban NeRF w/mask | 20.48 | 21.50 | 19.48 | 25.02 | 0.471 | 0.442 | 0.421 | 0.317 | 0.040 | 0.049 | 0.057 | 0.019 |
| Proposed | 20.64 | 23.63 | 25.18 | 31.74 | 0.190 | 0.161 | 0.192 | 0.084 | 0.023 | 0.011 | 0.008 | 0.005 |

In Figs. S3, S4, and S5 we show further renders from our method and baseline methods trained on the simulated dataset. The same trends as in the main text persist. Our method qualitatively produces images more faithful to the ground-truth. Our rendered images suffer less from floating artifacts and display finer details, for example in the *lego* scene. Some artifacts in the depth maps (i.e., the "holes" that appear) result from how we visualize depth in low occupancy areas. Depths in these regions can be become biased as described by Equation S6.

We show a breakdown across all simulated scenes of the evaluation metrics (see Table S7, S8, S9, S10). In addition to the metrics reported in the main text (PSNR, LPIPS, L1 depth) we add the structural similarity (SSIM) metric. Again, our method outperforms the baselines in the quantitative metrics. Since the 2, 3, and 5 view results do not include viewpoints that are strict supersets of each other, the performance of some metrics does not always increase for every scene in every case with increasing views (though this is a trend we observe on average).

Table S7: Breakdown of PSNR (dB) across all 5 simulated scenes.

| Scene | Instant NGP [7]↑ | | | DS-NeRF [9] | | | Urban NeRF [12] | | | Urban NeRF w/Mask [12] | | | Proposed | | |
|---|---|---|---|---|---|---|---|---|---|---|---|---|---|---|---|
| | 2 views | 3 views | 5 views | 2 views | 3 views | 5 views | 2 views | 3 views | 5 views | 2 views | 3 views | 5 views | 2 views | 3 views | 5 views |
| lego | 16.04 | 17.46 | 16.46 | 17.66 | 19.45 | 17.21 | 19.59 | 19.75 | 16.19 | 20.48 | 21.50 | 19.48 | 20.64 | 23.63 | 25.81 |
| chair | 13.40 | 15.42 | 24.95 | 16.51 | 16.72 | 25.49 | 17.53 | 16.14 | 24.28 | 20.39 | 18.69 | 28.23 | 20.75 | 21.99 | 34.48 |
| hotdog | 14.95 | 14.72 | 13.80 | 18.93 | 17.47 | 19.12 | 15.75 | 16.37 | 18.04 | 18.90 | 18.71 | 19.53 | 20.74 | 22.64 | 32.36 |
| bench | 16.85 | 19.33 | 15.58 | 19.87 | 20.34 | 15.69 | 19.12 | 19.52 | 14.56 | 20.64 | 21.55 | 17.57 | 20.20 | 23.06 | 21.57 |
| ficus | 21.87 | 21.40 | 27.50 | 23.44 | 22.75 | 27.85 | 22.32 | 21.88 | 25.91 | 24.12 | 23.58 | 26.87 | 24.57 | 26.10 | 27.70 |
| average | 16.62 | 17.67 | 19.66 | 19.28 | 19.35 | 21.07 | 18.86 | 18.73 | 19.80 | 20.91 | 20.81 | 22.34 | 21.38 | 23.48 | 28.39 |

Table S8: Breakdown of LPIPS metric across all 5 simulated scenes.

| Scene | Instant NGP [7]↑ | | | DS-NeRF [9] | | | Urban NeRF [12] | | | Urban NeRF w/Mask [12] | | | Proposed | | |
|---|---|---|---|---|---|---|---|---|---|---|---|---|---|---|---|
| | 2 views | 3 views | 5 views | 2 views | 3 views | 5 views | 2 views | 3 views | 5 views | 2 views | 3 views | 5 views | 2 views | 3 views | 5 views |
| lego | 0.591 | 0.544 | 0.479 | 0.500 | 0.476 | 0.482 | 0.519 | 0.522 | 0.503 | 0.471 | 0.442 | 0.421 | 0.190 | 0.161 | 0.192 |
| chair | 0.501 | 0.466 | 0.296 | 0.482 | 0.461 | 0.306 | 0.494 | 0.474 | 0.326 | 0.359 | 0.357 | 0.275 | 0.138 | 0.138 | 0.037 |
| hotdog | 0.583 | 0.615 | 0.523 | 0.456 | 0.470 | 0.390 | 0.580 | 0.553 | 0.433 | 0.533 | 0.466 | 0.395 | 0.242 | 0.241 | 0.118 |
| bench | 0.538 | 0.442 | 0.398 | 0.444 | 0.422 | 0.459 | 0.502 | 0.473 | 0.483 | 0.400 | 0.368 | 0.375 | 0.194 | 0.139 | 0.159 |
| ficus | 0.387 | 0.311 | 0.238 | 0.272 | 0.352 | 0.244 | 0.403 | 0.400 | 0.286 | 0.287 | 0.278 | 0.229 | 0.094 | 0.079 | 0.069 |
| average | 0.520 | 0.476 | 0.387 | 0.431 | 0.436 | 0.376 | 0.500 | 0.484 | 0.406 | 0.410 | 0.382 | 0.339 | 0.172 | 0.151 | 0.115 |

Table S9: Breakdown of SSIM metric across all 5 simulated scenes.

| Scene | Instant NGP [7]↑ | | | DS-NeRF [9] | | | Urban NeRF [12] | | | Urban NeRF w/Mask [12] | | | Proposed | | |
|---|---|---|---|---|---|---|---|---|---|---|---|---|---|---|---|
| | 2 views | 3 views | 5 views | 2 views | 3 views | 5 views | 2 views | 3 views | 5 views | 2 views | 3 views | 5 views | 2 views | 3 views | 5 views |
| lego | 0.456 | 0.519 | 0.576 | 0.600 | 0.624 | 0.631 | 0.592 | 0.580 | 0.590 | 0.646 | 0.710 | 0.722 | 0.860 | 0.897 | 0.899 |
| chair | 0.545 | 0.632 | 0.767 | 0.653 | 0.659 | 0.760 | 0.582 | 0.624 | 0.766 | 0.774 | 0.756 | 0.897 | 0.901 | 0.899 | 0.977 |
| hotdog | 0.508 | 0.448 | 0.534 | 0.736 | 0.636 | 0.687 | 0.503 | 0.534 | 0.664 | 0.567 | 0.686 | 0.761 | 0.882 | 0.875 | 0.963 |
| bench | 0.537 | 0.591 | 0.588 | 0.703 | 0.664 | 0.618 | 0.554 | 0.615 | 0.580 | 0.717 | 0.769 | 0.747 | 0.856 | 0.884 | 0.870 |
| ficus | 0.682 | 0.819 | 0.851 | 0.831 | 0.789 | 0.873 | 0.734 | 0.711 | 0.844 | 0.819 | 0.840 | 0.913 | 0.925 | 0.934 | 0.942 |
| average | 0.546 | 0.602 | 0.663 | 0.705 | 0.675 | 0.714 | 0.593 | 0.613 | 0.689 | 0.705 | 0.752 | 0.808 | 0.885 | 0.898 | 0.930 |

Table S10: Breakdown of L1 (depth) metric across all 5 simulated scenes.

| Scene | Instant NGP [7]↑ | | | DS-NeRF [9] | | | Urban NeRF [12] | | | Urban NeRF w/Mask [12] | | | Proposed | | |
|---|---|---|---|---|---|---|---|---|---|---|---|---|---|---|---|
| | 2 views | 3 views | 5 views | 2 views | 3 views | 5 views | 2 views | 3 views | 5 views | 2 views | 3 views | 5 views | 2 views | 3 views | 5 views |
| lego | 0.224 | 0.236 | 0.195 | 0.136 | 0.127 | 0.193 | 0.105 | 0.100 | 0.151 | 0.040 | 0.049 | 0.057 | 0.023 | 0.011 | 0.008 |
| chair | 0.166 | 0.198 | 0.117 | 0.123 | 0.146 | 0.097 | 0.130 | 0.132 | 0.066 | 0.034 | 0.057 | 0.009 | 0.006 | 0.005 | 0.003 |
| hotdog | 0.344 | 0.254 | 0.271 | 0.131 | 0.151 | 0.115 | 0.301 | 0.270 | 0.089 | 0.126 | 0.040 | 0.008 | 0.007 | 0.006 | 0.003 |
| bench | 0.166 | 0.137 | 0.198 | 0.082 | 0.076 | 0.142 | 0.060 | 0.047 | 0.168 | 0.011 | 0.007 | 0.038 | 0.006 | 0.005 | 0.003 |
| ficus | 0.291 | 0.147 | 0.107 | 0.072 | 0.074 | 0.048 | 0.059 | 0.073 | 0.034 | 0.043 | 0.034 | 0.032 | 0.034 | 0.033 | 0.048 |
| average | 0.238 | 0.195 | 0.178 | 0.109 | 0.115 | 0.119 | 0.131 | 0.124 | 0.101 | 0.051 | 0.038 | 0.029 | 0.015 | 0.011 | 0.013 |

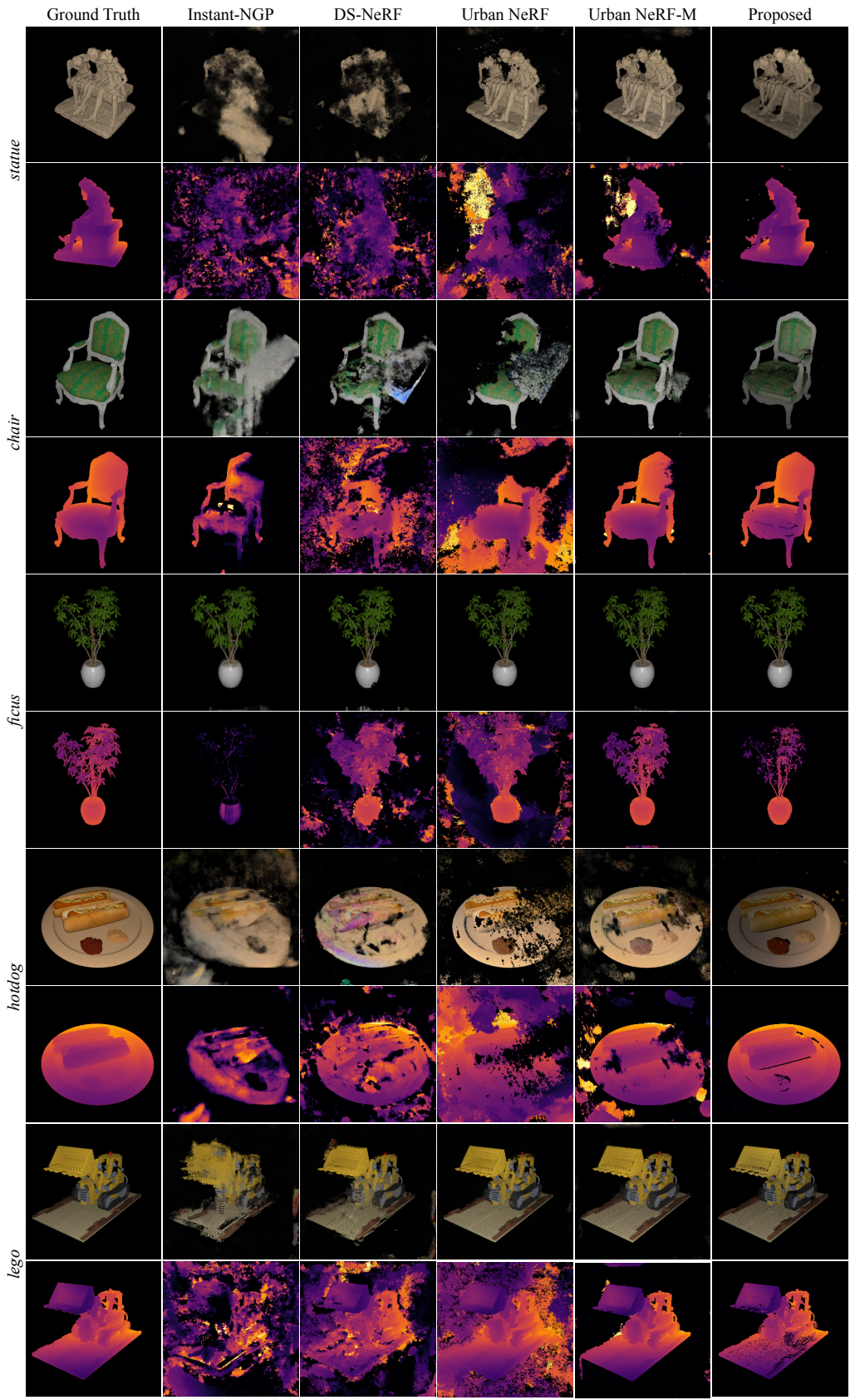

Figure S3: Rendered images and depths on the simulated dataset for 2 views.

11

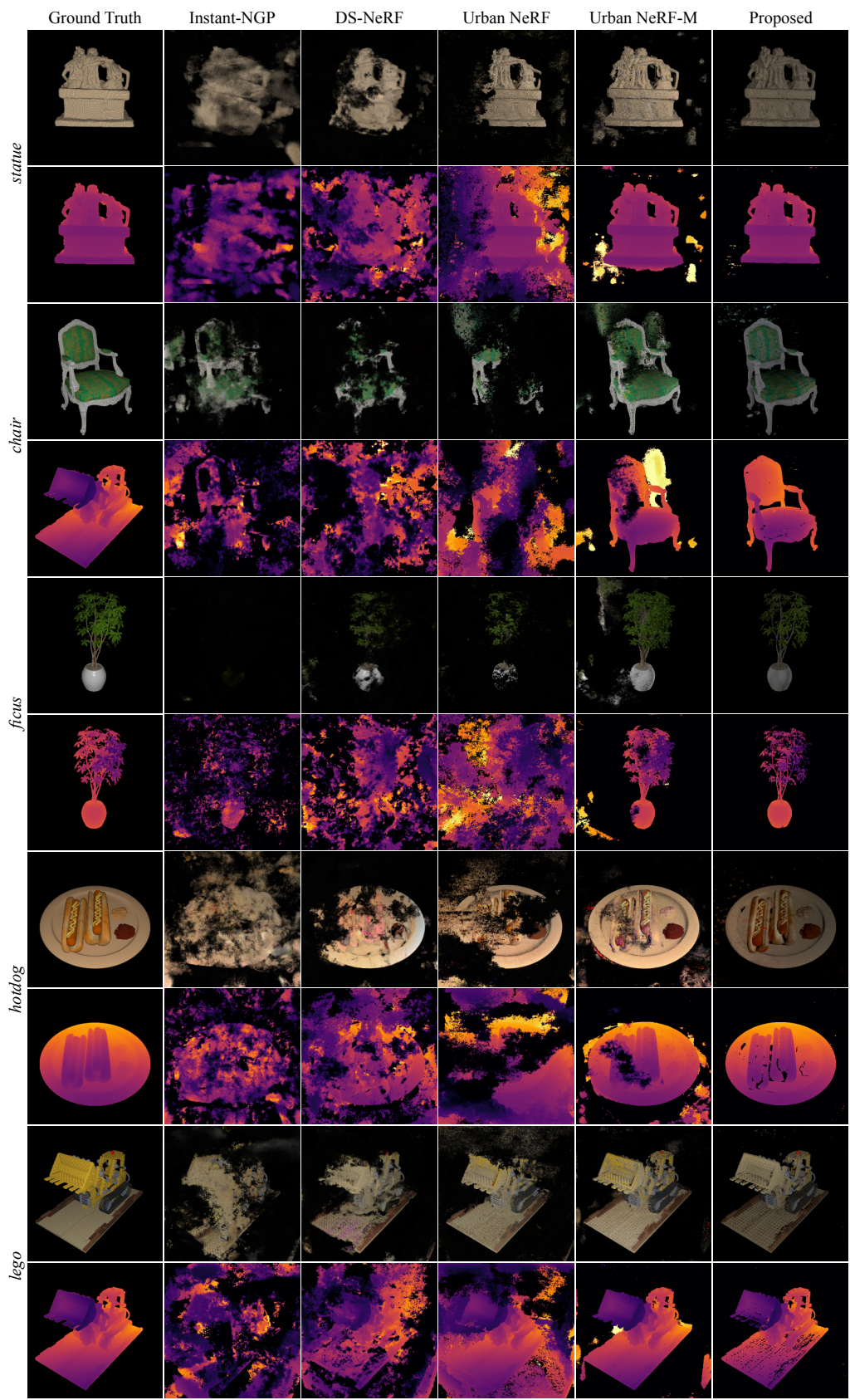

Figure S4: Rendered images and depths on the simulated dataset for 3 views.

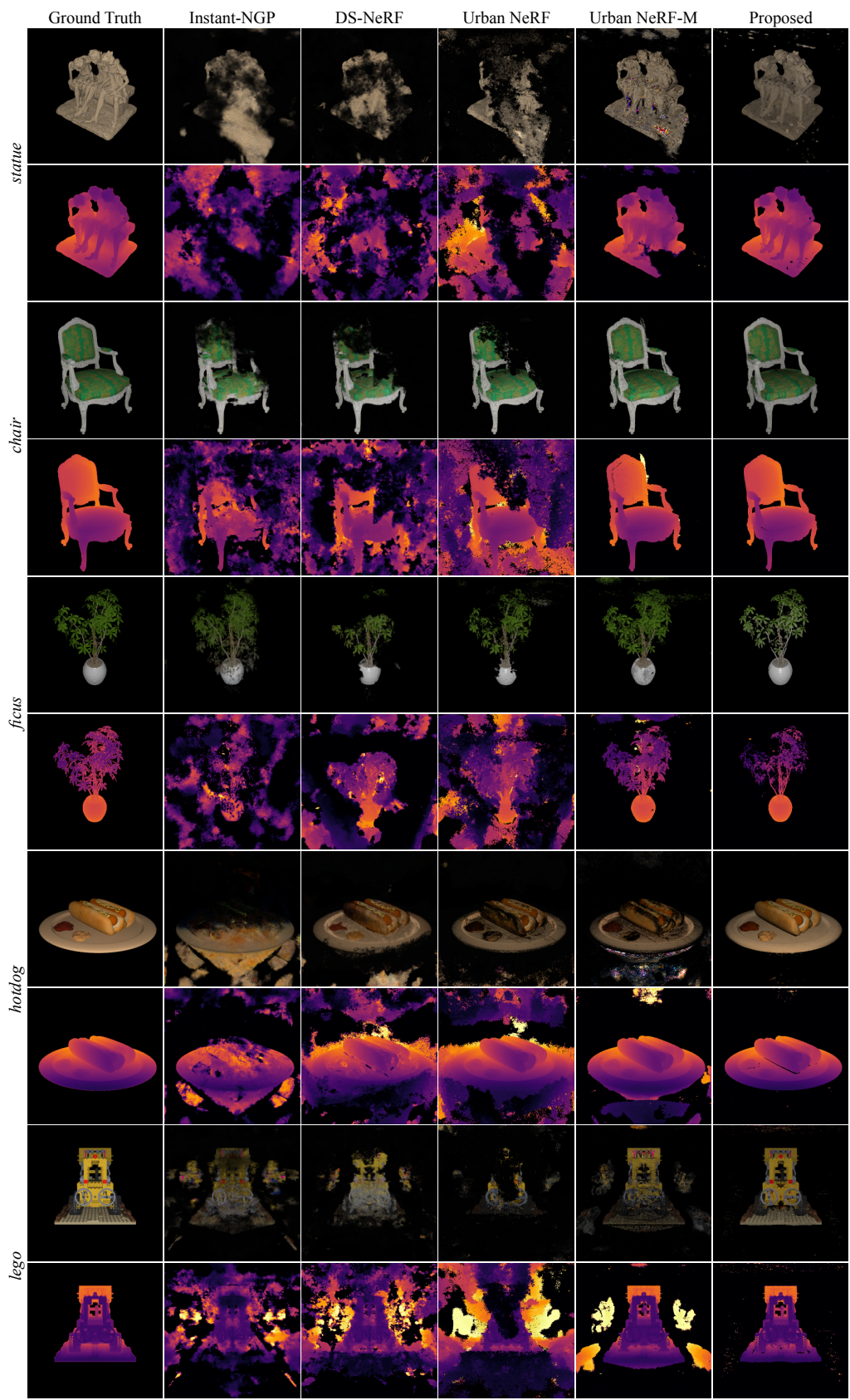

Figure S5: Rendered images and depths on the simulated dataset for 5 views.

### 4.6 Captured Results

In Figs. S6, S7, and S8 we show further renders from our model and the baselines on our captured dataset. The proposed method produces images more faithful to the ground truth.

We note that the quality of the reconstructions is lower than the simulation results. We attribute this to small imperfections in the calibration of the camera extrinsics, which register the multiview lidar scans to an accuracy of approximately 1 mm. Achieving precise, sub-mm alignment of the multiview lidar scans is a highly non-trivial problem and is outside the scope of our current work. The captured results bear out the trends observed in simulation and demonstrate Transient NeRF and novel view synthesis of lidar measurements for the first time in practice.

We show a breakdown across all captured scenes of the evaluation metrics (see Tables S11, S12, S13, and S14). In addition to the metrics reported in the main text (PSNR, LPIPS, L1 depth) we add the SSIM metric. Again our method outperforms the baselines in the quantitative metrics.

Table S11: Breakdown of PSNR (dB) across all 6 captured scenes.

| Scene | Instant NGP [7]↑ | | | DS-NeRF [9] | | | Urban NeRF [12] | | | Urban NeRF w/Mask [12] | | | Proposed | | |
|---|---|---|---|---|---|---|---|---|---|---|---|---|---|---|---|
| | 2 views | 3 views | 5 views | 2 views | 3 views | 5 views | 2 views | 3 views | 5 views | 2 views | 3 views | 5 views | 2 views | 3 views | 5 views |
| cinema | 14.73 | 15.18 | 15.21 | 13.18 | 13.87 | 12.58 | 17.27 | 15.11 | 17.30 | 14.13 | 17.22 | 17.80 | 21.61 | 21.66 | 25.12 |
| boots | 16.54 | 18.32 | 18.09 | 16.97 | 15.65 | 16.59 | 16.54 | 14.54 | 16.59 | 13.93 | 18.70 | 21.87 | 22.38 | 22.29 | 24.94 |
| baskets | 18.73 | 19.78 | 18.17 | 16.47 | 17.43 | 17.54 | 16.68 | 15.88 | 13.42 | 17.17 | 16.51 | 14.26 | 22.48 | 20.93 | 19.90 |
| carving | 17.96 | 17.30 | 16.70 | 17.41 | 16.36 | 15.44 | 18.71 | 17.97 | 17.36 | 16.07 | 21.41 | 22.28 | 23.52 | 24.20 | 24.70 |
| chef | 13.04 | 11.72 | 13.76 | 12.32 | 12.27 | 11.51 | 13.34 | 13.71 | 13.15 | 14.06 | 15.25 | 16.72 | 19.27 | 18.14 | 19.55 |
| food | 17.64 | 16.83 | 16.45 | 15.68 | 14.70 | 15.48 | 18.86 | 18.27 | 17.77 | 17.37 | 20.45 | 21.72 | 23.40 | 23.78 | 22.09 |
| average | 16.44 | 16.52 | 16.39 | 15.34 | 15.05 | 14.86 | 16.90 | 15.91 | 15.93 | 15.45 | 18.26 | 19.11 | 22.11 | 21.83 | 22.72 |

Table S12: Breakdown of LPIPS across all 6 captured scenes.

| Scene | Instant NGP [7]↑ | | | DS-NeRF [9] | | | Urban NeRF [12] | | | Urban NeRF w/Mask [12] | | | Proposed | | |
|---|---|---|---|---|---|---|---|---|---|---|---|---|---|---|---|
| | 2 views | 3 views | 5 views | 2 views | 3 views | 5 views | 2 views | 3 views | 5 views | 2 views | 3 views | 5 views | 2 views | 3 views | 5 views |
| cinema | 0.445 | 0.374 | 0.314 | 0.396 | 0.364 | 0.429 | 0.369 | 0.350 | 0.273 | 0.457 | 0.295 | 0.244 | 0.281 | 0.245 | 0.178 |
| boots | 0.251 | 0.253 | 0.181 | 0.193 | 0.225 | 0.192 | 0.386 | 0.332 | 0.163 | 0.525 | 0.245 | 0.135 | 0.221 | 0.182 | 0.155 |
| baskets | 0.399 | 0.298 | 0.252 | 0.256 | 0.244 | 0.268 | 0.444 | 0.311 | 0.220 | 0.433 | 0.300 | 0.195 | 0.269 | 0.165 | 0.164 |
| carving | 0.174 | 0.167 | 0.183 | 0.168 | 0.186 | 0.225 | 0.357 | 0.217 | 0.146 | 0.436 | 0.170 | 0.125 | 0.232 | 0.138 | 0.103 |
| chef | 0.580 | 0.443 | 0.401 | 0.498 | 0.434 | 0.467 | 0.513 | 0.468 | 0.362 | 0.473 | 0.373 | 0.270 | 0.334 | 0.338 | 0.276 |
| food | 0.299 | 0.310 | 0.313 | 0.354 | 0.417 | 0.369 | 0.351 | 0.290 | 0.221 | 0.425 | 0.232 | 0.174 | 0.286 | 0.205 | 0.154 |
| average | 0.358 | 0.307 | 0.274 | 0.311 | 0.312 | 0.325 | 0.403 | 0.328 | 0.231 | 0.458 | 0.269 | 0.191 | 0.271 | 0.212 | 0.172 |

Table S13: Breakdown of SSIM across all 6 captured scenes.

| Scene | Instant NGP [7]↑ | | | DS-NeRF [9] | | | Urban NeRF [12] | | | Urban NeRF w/Mask [12] | | | Proposed | | |
|---|---|---|---|---|---|---|---|---|---|---|---|---|---|---|---|
| | 2 views | 3 views | 5 views | 2 views | 3 views | 5 views | 2 views | 3 views | 5 views | 2 views | 3 views | 5 views | 2 views | 3 views | 5 views |
| cinema | 0.545 | 0.628 | 0.717 | 0.620 | 0.666 | 0.591 | 0.602 | 0.673 | 0.768 | 0.509 | 0.728 | 0.807 | 0.850 | 0.812 | 0.879 |
| boots | 0.804 | 0.810 | 0.866 | 0.853 | 0.822 | 0.851 | 0.587 | 0.687 | 0.871 | 0.414 | 0.777 | 0.894 | 0.912 | 0.909 | 0.914 |
| baskets | 0.555 | 0.715 | 0.749 | 0.737 | 0.750 | 0.738 | 0.499 | 0.689 | 0.751 | 0.517 | 0.691 | 0.771 | 0.846 | 0.850 | 0.826 |
| carving | 0.852 | 0.853 | 0.843 | 0.858 | 0.841 | 0.804 | 0.620 | 0.827 | 0.877 | 0.526 | 0.854 | 0.906 | 0.913 | 0.929 | 0.929 |
| chef | 0.414 | 0.586 | 0.647 | 0.544 | 0.608 | 0.574 | 0.450 | 0.559 | 0.686 | 0.488 | 0.651 | 0.780 | 0.775 | 0.747 | 0.811 |
| food | 0.731 | 0.720 | 0.733 | 0.677 | 0.586 | 0.671 | 0.654 | 0.737 | 0.808 | 0.528 | 0.775 | 0.855 | 0.826 | 0.879 | 0.873 |
| average | 0.650 | 0.719 | 0.759 | 0.715 | 0.712 | 0.705 | 0.569 | 0.695 | 0.793 | 0.497 | 0.746 | 0.836 | 0.854 | 0.854 | 0.872 |

Table S14: Breakdown of L1 (depth) metric across all 6 captured scenes.

| Scene | Instant NGP [7]↑ | | | DS-NeRF [9] | | | Urban NeRF [12] | | | Urban NeRF w/Mask [12] | | | Proposed | | |
|---|---|---|---|---|---|---|---|---|---|---|---|---|---|---|---|
| | 2 views | 3 views | 5 views | 2 views | 3 views | 5 views | 2 views | 3 views | 5 views | 2 views | 3 views | 5 views | 2 views | 3 views | 5 views |
| cinema | 0.141 | 0.052 | 0.041 | 0.067 | 0.040 | 0.050 | 0.023 | 0.017 | 0.015 | 0.018 | 0.005 | 0.006 | 0.006 | 0.006 | 0.007 |
| boots | 0.034 | 0.076 | 0.048 | 0.029 | 0.023 | 0.022 | 0.017 | 0.014 | 0.011 | 0.013 | 0.003 | 0.003 | 0.001 | 0.002 | 0.002 |
| baskets | 0.138 | 0.141 | 0.040 | 0.038 | 0.033 | 0.029 | 0.015 | 0.017 | 0.004 | 0.016 | 0.008 | 0.003 | 0.008 | 0.008 | 0.026 |
| carving | 0.029 | 0.032 | 0.031 | 0.031 | 0.024 | 0.026 | 0.008 | 0.011 | 0.014 | 0.009 | 0.004 | 0.003 | 0.006 | 0.005 | 0.006 |
| chef | 0.199 | 0.100 | 0.124 | 0.062 | 0.056 | 0.055 | 0.023 | 0.020 | 0.025 | 0.017 | 0.011 | 0.009 | 0.003 | 0.006 | 0.006 |
| food | 0.146 | 0.057 | 0.035 | 0.058 | 0.041 | 0.035 | 0.017 | 0.013 | 0.014 | 0.013 | 0.006 | 0.005 | 0.006 | 0.007 | 0.016 |
| average | 0.115 | 0.076 | 0.053 | 0.048 | 0.036 | 0.036 | 0.017 | 0.015 | 0.014 | 0.014 | 0.006 | 0.005 | 0.005 | 0.006 | 0.010 |

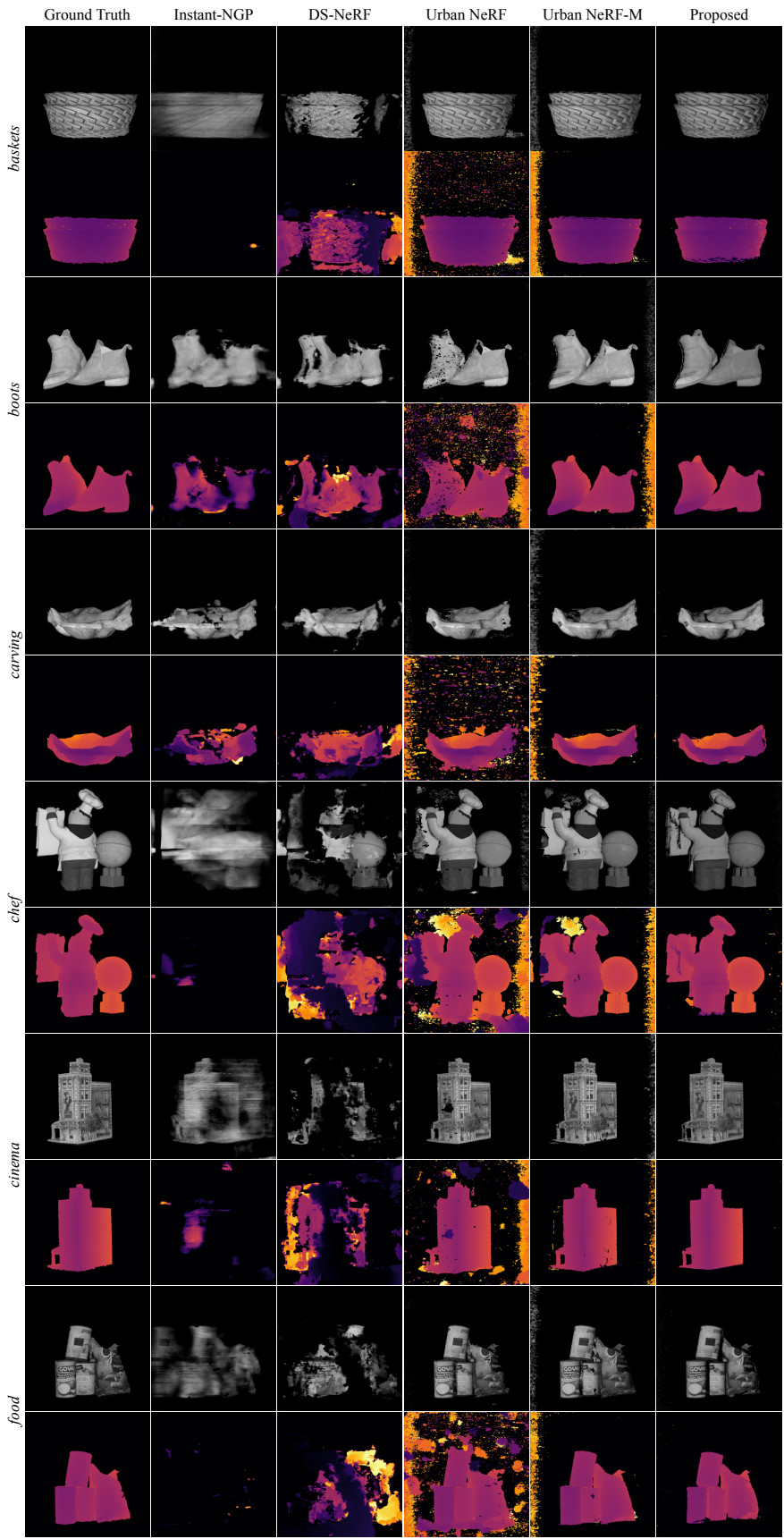

Figure S6: Rendered images and depths on the captured dataset for 2 views.

15

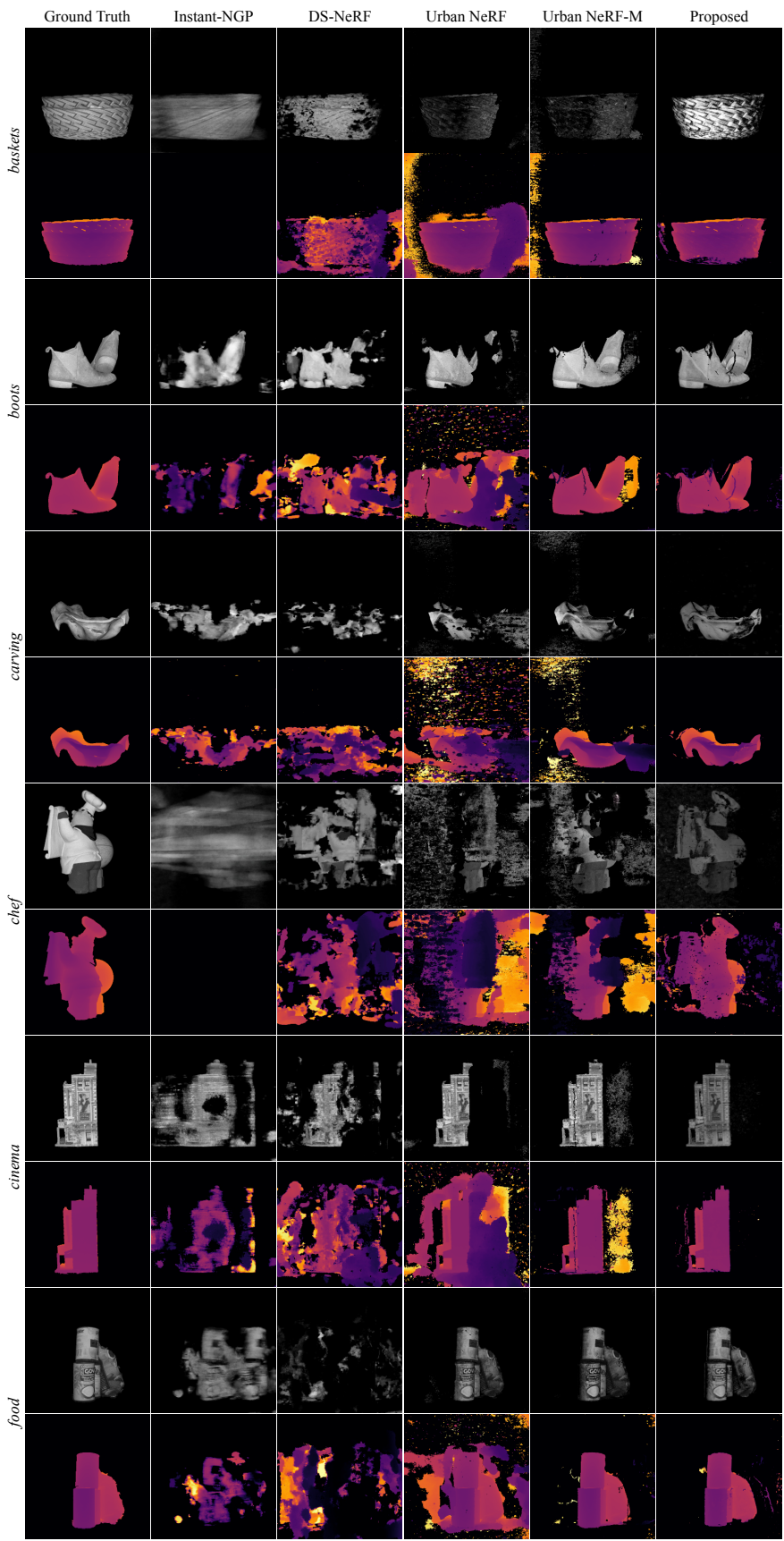

Figure S7: Rendered images and depths on the captured dataset for 3 views.

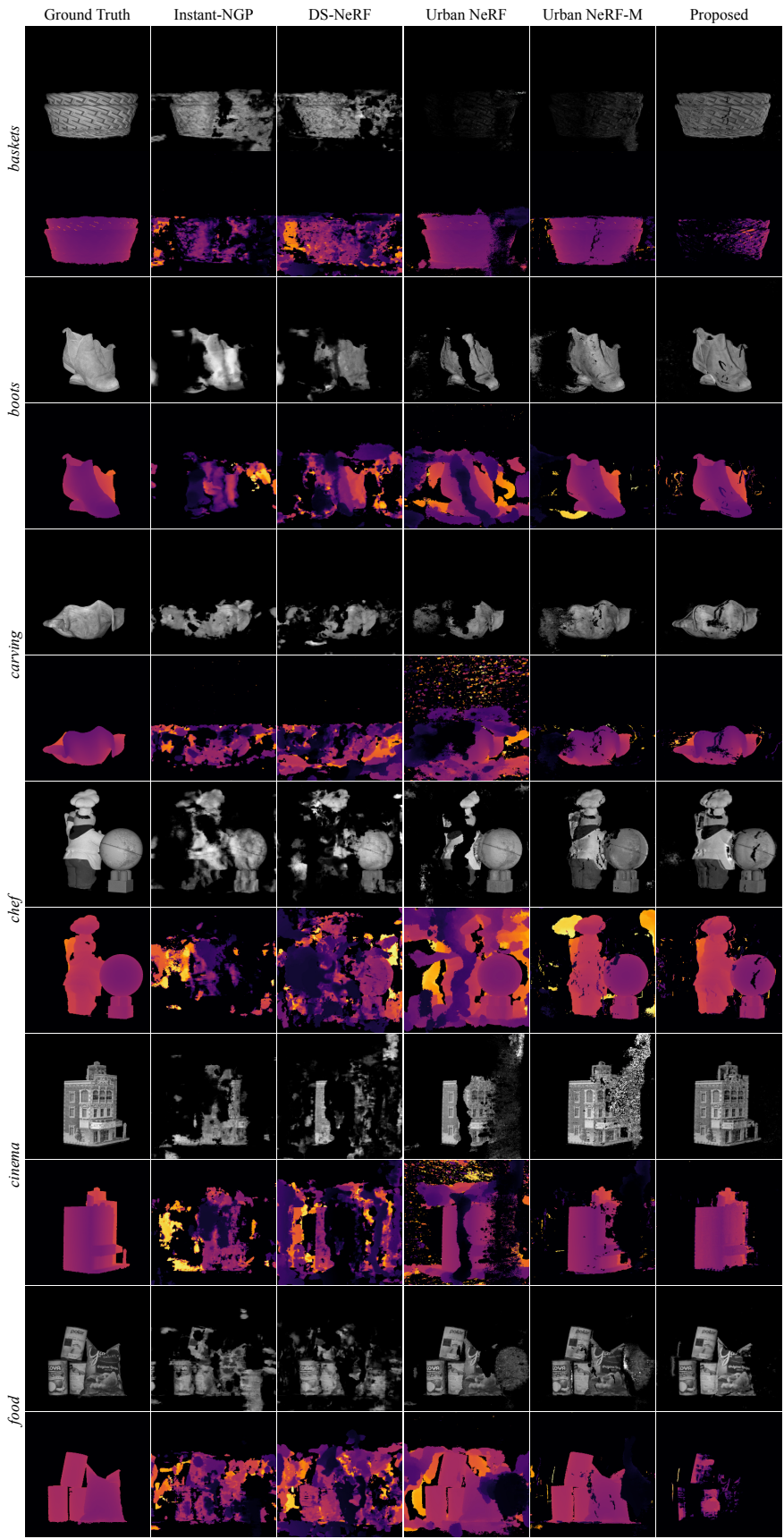

Figure S8: Rendered images and depths on the captured dataset for 5 views.



\end{document}